\documentclass[10pt,twocolumn,letterpaper]{article}

\usepackage{cvpr}              

\usepackage{graphicx}
\usepackage{amsmath}
\usepackage{booktabs}
\usepackage{float}

\usepackage[pagebackref,breaklinks,colorlinks]{hyperref}

\usepackage[capitalize]{cleveref}
\crefname{section}{Sec.}{Secs.}
\crefname{table}{Tab.}{Tabs.}

\def\cvprPaperID{5457} 
\def\confName{CVPR}
\def\confYear{2022}

\graphicspath{{images/}{images/vilt_ot_heatmap}}

\begin{document}

\title{Winoground: Probing Vision and Language Models\\ for Visio-Linguistic Compositionality} 

\newcommand*\samethanks[1][\value{footnote}]{\footnotemark[#1]}
\author{Tristan Thrush$^\mathparagraph$\thanks{Equal contribution. TT, AS, and DK conducted most of the work for this paper when they were at Facebook AI Research.}, Ryan Jiang$^\ddagger$, Max Bartolo$^\mathsection$,\\ Amanpreet Singh$^\mathparagraph$, Adina Williams$^\dagger$, Douwe Kiela$^\mathparagraph$, Candace Ross$^\dagger$\samethanks\\
$^\mathparagraph$ Hugging Face; $^\dagger$ Facebook AI Research; $^\ddagger$ University of Waterloo; $^\mathsection$ University College London\\
{\tt\small tristan@huggingface.co, ccross@fb.com}
}
\maketitle


\begin{abstract}
We present a novel task and dataset for evaluating the ability of vision and language models to 
conduct visio-linguistic compositional reasoning, which we call Winoground. Given two images and two captions, the goal is to match them correctly---but crucially, both captions contain a completely identical set of words, 
only in a different order. The dataset was carefully hand-curated by expert annotators and is labeled with a rich set of fine-grained tags to assist in analyzing model performance. We probe a diverse range of state-of-the-art vision and language models and find that, surprisingly, none of them do much better than chance. Evidently, these models are not as skilled at visio-linguistic compositional reasoning as we might have hoped. We perform an extensive analysis to obtain insights into how future work might try to mitigate these models' shortcomings. We aim for Winoground to serve as a useful evaluation set for advancing the state of the art and driving further progress in the field. The dataset is available at \scriptsize{\url{https://huggingface.co/datasets/facebook/winoground}}.
\end{abstract}

\section{Introduction}
\label{sec:intro}

Despite the impressive performance of pretrained vision and language transformers on a wide variety of multimodal tasks~\cite{lu2019vilbert, li2019visualbert, radford2021clip}, they remain poorly understood~\cite{dou2021empirical,cao2020behind,li2020closer,singh2020we}. One important question is to what extent such models are able to conduct unimodal and multimodal compositional reasoning. For humans, the visual differences between images depicting ``the tree is in the shopping cart’’ and ``the shopping cart is in the tree’’ will be blatantly obvious, even when the words in the captions are identical---but is the same true for machines?

While matching simple images and captions may seem almost too trivial a task, recent work in NLP has shown that transformers are often remarkably insensitive to word order \cite{sinha2021matterslittle}. Understanding the relationship between text in captions and corresponding visual content is a fundamental goal of computer vision, and the fact that different word orders correspond to wildly different visual depictions should be reflected in the capabilities of our models.

Motivated by this, we propose a novel task, called Winoground, for measuring visio-linguistic compositional reasoning, 
whereby two images and two captions have to be matched correctly; both captions contain exactly the same set of words, ordered in such a way that each describes primarily one of the images. To perform well on Winoground, models must not only encode text and images well (i.e., be sensitive to the compositional structure present in each modality), but they also must be able to synthesize information across the two modalities.

\begin{figure}[!t]%
    \centering
    \subfloat[\centering some plants surrounding a lightbulb]{{\includegraphics[height=3cm]{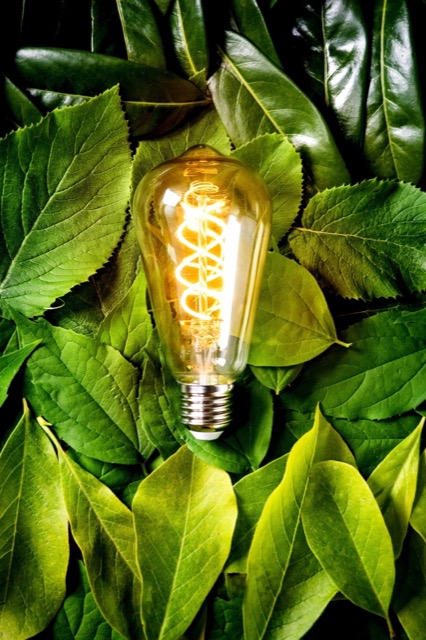} }}%
    \qquad
    \subfloat[\centering a lightbulb surrounding some plants]{{\includegraphics[height=3cm]{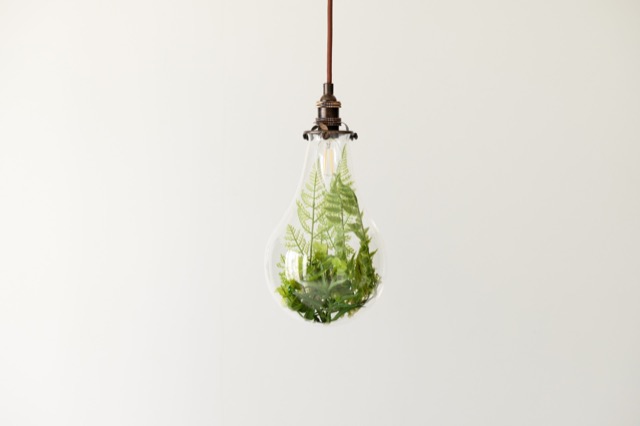} }}%
    \caption{An example from Winoground. The two sentences contain the same words but in a different order. The task of understanding which image and caption match is trivial for humans but much harder for vision and language models. Every model that we tested (UNITER, ViLLA, VinVL, VisualBERT, ViLT, LXMERT, ViLBERT, UniT, FLAVA, CLIP, VSE++, and VSRN) fails to correctly pair the images and captions, except the large checkpoint of ViLLA by a very thin margin (0.00013 confidence).}%
    \label{fig:teaser-ex}%
\end{figure}

We draw inspiration from the Winograd Schema Challenge \cite{levesque2012winograd}, which tests the commonsense capabilities of models. In the challenge, a model is given two sentences that minimally differ and is tasked with performing coreference resolution. The Winograd twin sentence format has been used for a variety of language-related tasks \cite{rudinger2018gender, zhao2018gender, sakaguchi2020winogrande}.
In this work, we study the image-grounding of twin sentences with identical but differently ordered words.

Winoground was hand-crafted by expert annotators and is labeled with a rich set of fine-grained tags to assist in analyzing model performance. In efforts to shed better light on what exactly models learn, the NLP community has designed a wide variety of ``probing tasks’’: specialized, targeted tasks meant specifically for evaluation. The primary purpose of Winoground is to serve as a probing task for vision and language models. See \cref{fig:teaser-ex} for an example.

We evaluate a variety of state-of-the-art vision and language (V\&L) transformers \cite{gan2020villa,li2019visualbert,lu2019vilbert,hu2021unit,chen2020uniter,tan2020lxmert,radford2021clip,zhang2021vinvl,kim2021vilt,singh2022flava} and RNN-based models \cite{faghri2018vse,li2019vsrn}. Surprisingly,
all of the models rarely---and if so only barely---outperform chance. 
Our findings indicate that the visio-linguistic compositional reasoning capabilities of these models fall dramatically short of what we might have hoped.

In what follows, we introduce the Winoground task and dataset. We then describe the models we tested and discuss our findings. Next, we conduct an analysis of the performance of different models. We hope that insights from this work will lead to more robust vision and language models.

\section{Related Work}


%
%
%

\textbf{Visio-linguistic stress testing.} There are a number of existing multimodal stress tests about correctly understanding implausible scenes \cite{choi2012context}, exploitation of language and vision priors \cite{chao2017being, goyal2017making}, single word mismatches \cite{shekhar2017foil}, hate speech detection \cite{hosseinmardi2015detection,zhong2016content,gomez2020exploring, kirk2021hatemoji}, memes \cite{kiela2020hateful,suryawanshi2021trollmeme}, ablation of one modality to probe the other \cite{frank-etal-2021-vision}, distracting models with visual similarity between images \cite{hu2019evaluating, bogin2021covr}, distracting models with textual similarity between many suitable captions \cite{ding2016understanding, akula2020words}, collecting more diverse image-caption pairs beyond the predominately English and North American/Western European datasets \cite{liu-etal-2021-visually}, probing for an understanding of verb-argument relationships \cite{hendricks2021probing}, counting \cite{parcalabescu2021seeing}, or specific model failure modes \cite{singh2019towards,sidorov2020textcaps}. Many of these stress tests rely only on synthetically generated images, often with minimal visual differences, but no correspondingly minimal textual changes \cite{vedantam2021curi}. Other datasets test models with a single caption \cite{suhr2017corpus} or a single image \cite{johnson2017clevr, bitton-etal-2021-automatic}. There are also purely visual stress tests with naturalistic images: ImageNet-C/ImageNet-P \cite{hendrycks2019robustness} tests models on perturbations for a variety of image features.  Unlike Winoground, these stress tests tend to come from existing datasets that have images and text from typical training domains, such as Conceptual Captions \cite{sharma2018conceptual}, COCO \cite{lin2014microsoft}, Visual7W \cite{zhu2016visual7w} and VQA \cite{antol2015vqa,goyal2017making}. None of them hold the set of words constant in the captions, which is what allows us to carefully test for compositional reasoning without any biases stemming from the presence of altogether different words. While it is theoretically possible for unstructured bag of words models to do well on these previous datasets, that is not possible on Winoground.

\textbf{Probing.} Measuring what exactly a model knows about word order and linguistic structure has been explored in natural language processing. Sinha et al. \cite{sinha2021matterslittle} found that word order information does not have a large impact on performance when pretraining large transformer language models, across a variety of metrics. This suggests that transformers use high-level word co-occurence statistics, which gives the illusion of an understanding of word order. Other work in this space has tried to understand what models know about syntax \cite{linzen2016assessing, gulordava2018, williams-etal-2018-latent, hu-etal-2020-systematic, gauthier-etal-2020-syntaxgym,  sinha-etal-2021-unnatural, parthasarathi-etal-2021-sometimes-want} or the complex interaction between syntactic and semantic categories \cite{kann-etal-2019-verb,warstadt-etal-2019-investigating,thrush2020investigating, warstadt-etal-2020-blimp-benchmark}.

\textbf{Winograd schemas.} The Winograd Schema Challenge~\cite{levesque2012winograd} was named after a coreference resolution problem presented by Terry Winograd \cite{winograd1972understanding}. The goal is to correctly resolve (an) ambiguous referent(s) in two English sentences. The sentences have a minor difference that changes how a human resolves the referent. Winograd schema examples are easily handled by humans, and commonsense reasoning is said to be required \cite{bender2015establishing}.
For example, in the sentence \textit{``The city councilmen refused the demonstrators a permit because they [feared/advocated] violence"}, the pronoun \textit{they} can either refer to the councilmen or to the demonstrators depending on which word is chosen.
The format has been used in a variety of other tasks and datasets. For instance, Sakaguchi et al.\cite{sakaguchi2020winogrande} introduce WinoGrande: a large-scale approach to building a Winograd Schema dataset that uses Amazon Mechanical Turk to generate sentences instead of expert annotators like the original work of Levesque et al. \cite{levesque2012winograd}.
Other approaches use ambiguous pronouns in sentences to probe for gender biases in models \cite{rudinger2018gender,zhao2018gender}.
See Kotcijan et al. \cite{kocijan2020review} for an in-depth review. Winoground is the first work to apply these ideas to the vision and language domain, by using twin captions with identical word content and two images that are each associated with one caption over the other. 
%
%

%




\section{Winoground}

In this section, we describe how the dataset was constructed and how performance on the task is to be measured.

\subsection{Dataset}
\begin{figure}
    \centering
        \begin{minipage}[t]{.15\textwidth}
            \begin{subfigure}[t]{\textwidth}
            \centering
            \includegraphics[height=1.85cm]{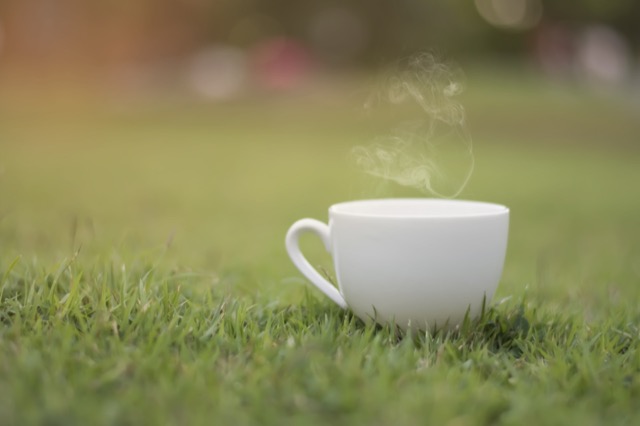}
            \caption{there is [a mug] in [some grass]}
            \end{subfigure}\\
            \begin{subfigure}[t]{\textwidth}
            \centering
            \includegraphics[height=1.85cm]{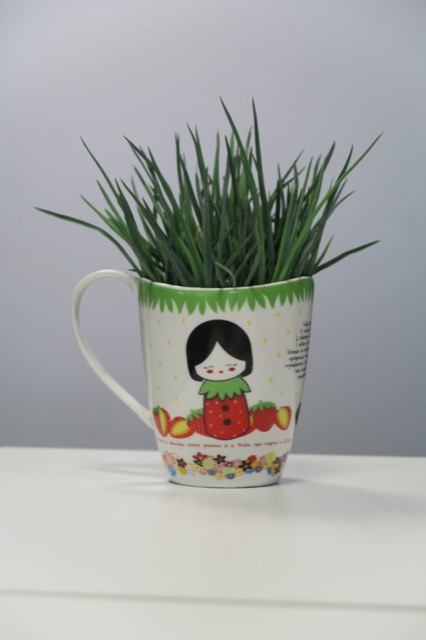}
            \caption{there is [some grass] in [a mug]}
            \end{subfigure}%
            \caption*{\textit{Object}}
        \end{minipage}
        \hfill
        \begin{minipage}[t]{.15\textwidth}
            \begin{subfigure}[t]{\textwidth}
            \centering
            \includegraphics[height=1.85cm]{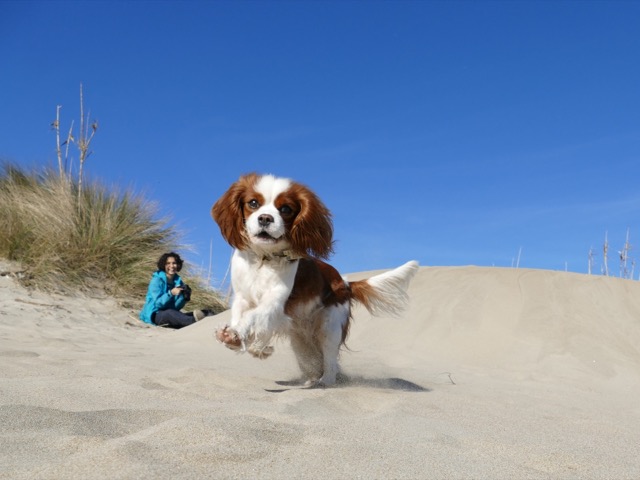}
            \caption{a person [sits] and a dog [stands]}
            \end{subfigure}\\
            \begin{subfigure}[t]{\textwidth}
            \centering
            \includegraphics[height=1.85cm]{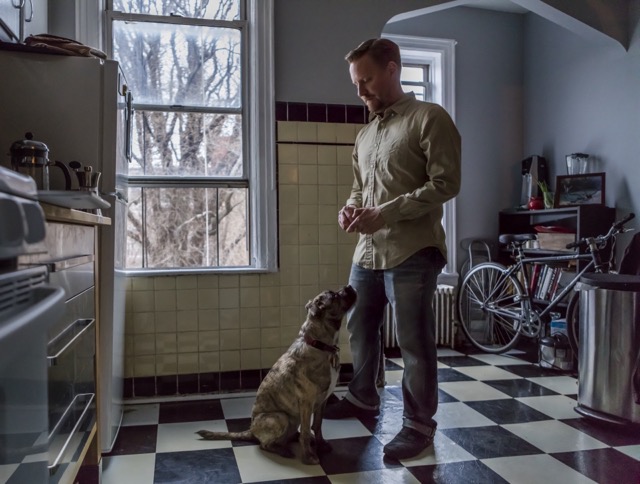}
            \caption{a person [stands] and a dog [sits]}
            \end{subfigure}%
            \caption*{\textit{Relation}}
        \end{minipage}
        \hfill
        \begin{minipage}[t]{.15\textwidth}
            \begin{subfigure}[t]{\textwidth}
            \centering
            \includegraphics[height=1.85cm]{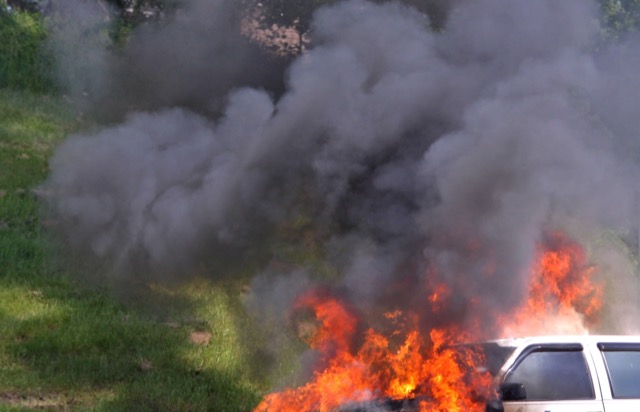}
            \caption{it's a [truck] [fire]}
            \end{subfigure}\\
            \vspace{9pt}
            \begin{subfigure}[t]{\textwidth}
            \centering
            \includegraphics[height=1.85cm]{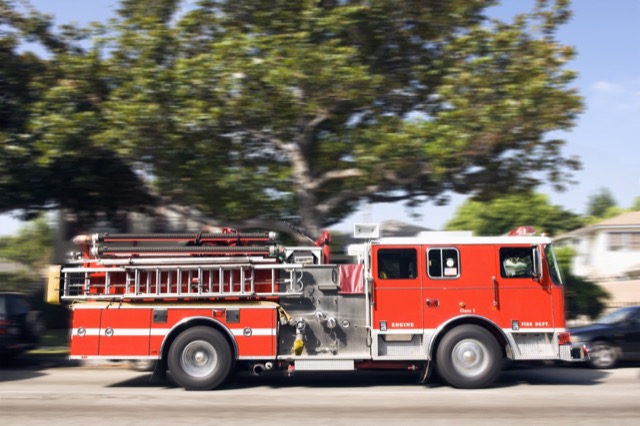}
            \caption{it's a [fire] [truck]}
            \end{subfigure}%
            \vspace{9pt}
            \caption*{\textit{Both}}
        \end{minipage}%
    \end{figure}
    
\begin{figure}
    \centering
        \begin{minipage}{.15\textwidth}
            \begin{subfigure}{\textwidth}
            \centering
            \includegraphics[height=1.85cm]{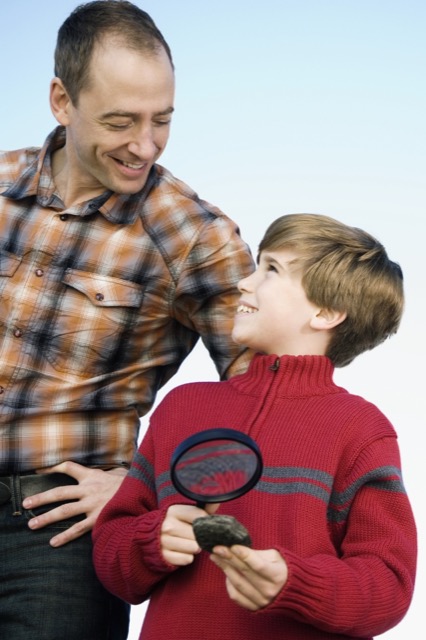}
            \caption{the kid [with the magnifying glass] looks at them []}
            \end{subfigure}\\
            \begin{subfigure}{\textwidth}
            \centering
            \includegraphics[height=1.85cm]{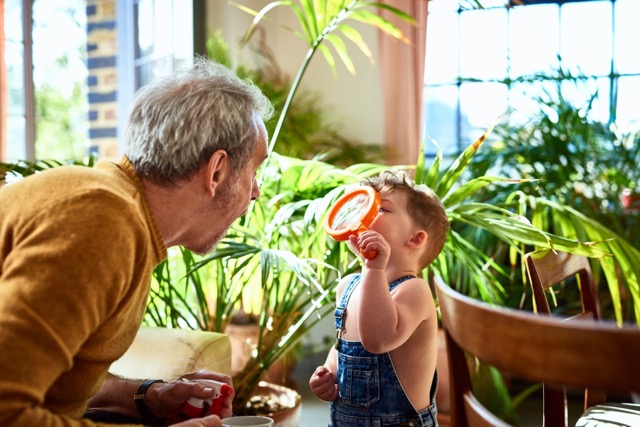}
            \caption{the kid [] looks at them [with the magnifying glass]}
            \end{subfigure}%
            \caption*{\textit{Pragmatics}}
        \end{minipage}
        \hfill
        \begin{minipage}{.15\textwidth}
            \begin{subfigure}{\textwidth}
            \centering
            \includegraphics[height=1.85cm]{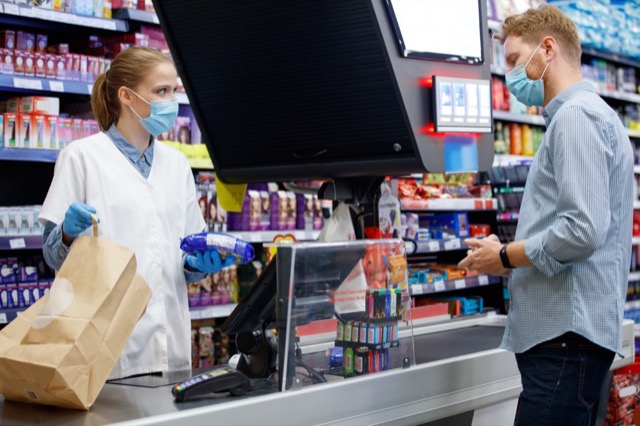}
            \caption{the person with the ponytail [packs] stuff and other [buys] it}
            \end{subfigure}\\
            \begin{subfigure}{\textwidth}
            \centering
            \includegraphics[height=1.85cm]{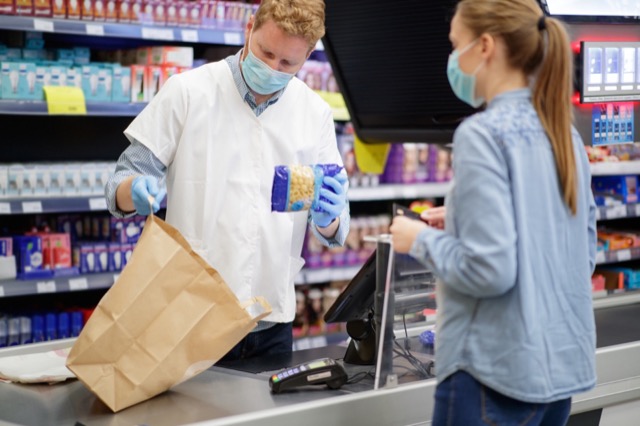}
            \caption{the person with the ponytail [buys] stuff and other [packs] it}
            \end{subfigure}%
            \caption*{\textit{Series}}
        \end{minipage}
        \hfill
        \begin{minipage}{.15\textwidth}
            \begin{subfigure}{\textwidth}
            \centering
            \includegraphics[height=1.85cm]{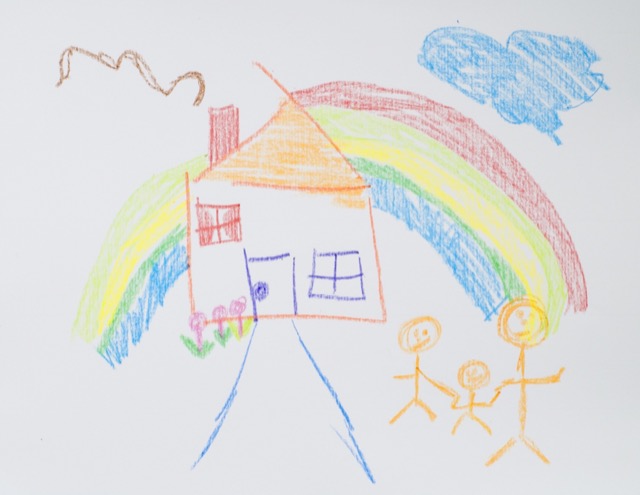}
            \caption{there are [three] people and [two] windows}
            \end{subfigure}\\
            \begin{subfigure}{\textwidth}
            \centering
            \includegraphics[height=1.85cm]{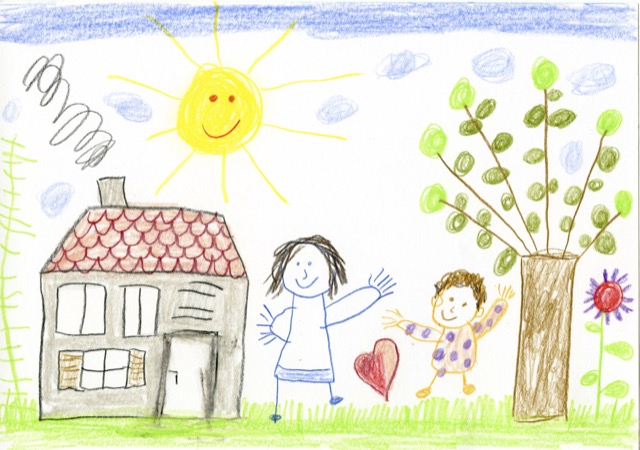}
            \caption{there are [two] people and [three] windows}
            \end{subfigure}%
            \caption*{\textit{Symbolic}}
        \end{minipage}
        \caption[]{Examples from our dataset for the swap-dependent linguistic tags (top) and visual tags (bottom). The visual examples are additionally tagged with the \textit{Relation} tag, and 1, 2, and 1 main predicates from left to right. The linguistic examples are additionally tagged with 2, 1, and 1 main predicates from left to right.}
        \label{fig:dataset-examples}
    \end{figure}

\begin{table}
\centering
\begin{tabular}{lrr}
\toprule
 Category & Tag    &   Count \\
\midrule
 & Object   &     141 \\
 Linguistic$_\text{swap-dep.}$ & Relation &     233 \\
 & Both &      26 \\\midrule
 Linguistic$_\text{swap-indep.}$ & 1 Main Pred & 293 \\
 & 2 Main Preds & 108 \\\midrule
 & Symbolic &  41 \\
 Visual & Series &  31 \\
 & Pragmatics &  24\\
\bottomrule
\end{tabular}
\caption{Linguistic and visual tag counts in the Winoground dataset. Every example has a linguistic tag; only examples that contain the visual phenomena have visual tags.}
\label{tab:stats-tag-subset}
\end{table}
%

The Winoground dataset was hand-curated by four expert annotators with extensive experience in vision and language research as well as computational linguistics. Let~$(C_{0},I_{0})$ and $(C_{1},I_{1})$ be two image-caption pairs. An example satisfies the Winoground schema if and only if:
\begin{itemize}
    \item $(C_{0},I_{0})$ and $(C_{1},I_{1})$ are preferred by the annotator over $(C_{1},I_{0})$ and $(C_{0},I_{1})$; and
    \item $C_{0}$ and $C_{1}$ have the same words and/or morphemes but the order differs.
\end{itemize}

We have secured a license from Getty Images to distribute images for research purposes. Thus, the expert annotators were given access to the Getty Images API \cite{getty}, and tasked with jointly creating captions and finding images to compose examples. We encouraged them to be as creative as possible, and to mark each of their examples with fine-grained linguistic tags. If applicable, annotators also marked examples with one or more visual reasoning tags.

The annotators created a total of 70 linguistic tags for the swaps that make caption pairs different. This set of tags can be split into three broad groups: objects, relations, and swaps involving both relations and objects. Object swaps reorder elements such as noun phrases that tend to refer to objects in the real world. Relation swaps reorder elements such as verbs, adjectives, prepositions, and/or adverbs, which tend to take nouns referring to objects as semantic arguments \cite{altshuler2019course}. Swaps of both relations and objects can involve two separate swaps, or can involve a single swap that changes parts of speech (e.g., ``it's a [fire] [truck]" vs. ``it's a [truck] [fire]"). Examples of each broad tag group can be seen in \cref{fig:dataset-examples}. For examples for each fine-grained linguistic tag, see Appendix~C.

Separately, the annotators tagged examples for how many main predicates were in the captions, which is not dependent on the specific swap happening between the two captions. For example, ``left is blue and right is red'' has two main predicates and ``water is in a bottle'' has one main predicate. It turned out that all examples in Winoground have either one main predicate or two.

Finally, examples were tagged from a set of three non-mutually exclusive visual reasoning tags, which are tied in some way to the images in an example, and not necessarily the captions. The ``Pragmatics'' tag comprises examples where the images need to be interpreted non-literally due to idiomatic uses of language in a caption (e.g. ``it starts with Z and ends with A'' describing an image of a Zebra) or due to attachment preferences of prepositional phrases in the captions (e.g. ``the kid looks at them with the magnifying glass'' describing an image of a child looking at someone through a magnifying glass with greater confidence than an image of a child looking at someone while holding a magnifying glass at their side). The ``Symbolic" tag represents whether a symbolic depiction of something must be understood to make a correct prediction (e.g., objects in a child's drawing). Lastly, the ``Series'' tag is given to examples where both images come from the same photo series on Getty, which typically means that the same people occur in both images, with a similar background and in similar lighting.

See \cref{fig:dataset-examples} for representative examples of the tags, and \cref{tab:stats-tag-subset} for tag counts. As noted, Winoground is a probing dataset and so we prioritize clean, expert annotations over mere size. Our dataset has 1600 image-text pairs in total, with 800 correct and 800 incorrect pairings. These comprise 400 examples, with 800 unique captions and images.




\subsection{Metrics}\label{sec:metrics}
Performance on Winoground is computed according to three different metrics that evaluate different aspects of the models' visio-linguistic reasoning abilities.
The first metric is the \textbf{text score}, which measures whether a model can select the correct caption, given an image.
Given images $I_0$ and $I_{1}$ and captions $C_{0}$ and $C_{1}$, the text score for an example $(C_{0},I_{0},C_{1},I_{1})$ is computed according to:
\begin{equation}\label{eq:text-score}
        f(C_{0},I_{0},C_{1},I_{1})= 
    \begin{cases}
        1 & \text{if}\  s(C_{0}, I_{0}) > s(C_{1}, I_{0}) \\
        & \ \ \text{and}\ s(C_{1}, I_{1}) > s(C_{0}, I_{1}) \\
        0              & \text{otherwise}
    \end{cases}
\end{equation}
where $s(\cdot)$ is the model's score for the image/caption pair.
This metric tests whether the ground truth caption for a given image in our dataset is scored higher than the alternative caption \textit{and} whether this holds for the other image/caption pair in the example too.

The second metric is the \textbf{image score}, which measures whether a model can select the correct image, given a caption.
Given images $I_0$ and $I_{1}$ and captions $C_{0}$ and $C_{1}$, the image score for an example is computed according to:
\begin{equation}\label{eq:image-score}
        g(C_{0},I_{0},C_{1},I_{1})= 
    \begin{cases}
        1 & \text{if}\  s(C_{0}, I_{0}) > s(C_{0}, I_{1})\\
        & \ \ \text{and}\ s(C_{1}, I_{1}) > s(C_{1}, I_{0}) \\
        0              & \text{otherwise}
    \end{cases}
\end{equation}
This metric tests whether the ground truth image for a given caption is scored higher than the image corresponding to the alternative caption \textit{and} whether this holds vice versa.

Our final metric combines the previous two. In their analysis of the Winograd Schema Challenge, Elazar et al.~\cite{elazar2021back} find that evaluation metrics tend to overestimate model performance by computing scores for the twin sentences individually instead of as a set. So, we also evaluate using the \textbf{group score}, where every combination for a given example \{$(C_{0}, I_{0}), (C_{0}, I_{1}), (C_{1}, I_{0}), (C_{1}, I_{1})$\} must be correctly scored by the model in order for the example to be considered correct.
The group score in our framework is computed according to:
\begin{equation}\label{eq:group-score}
        h(C_{0},I_{0},C_{1},I_{1})= 
    \begin{cases}
        1 & \text{if}\  f(C_{0},I_{0},C_{1},I_{1})  \\
         & \ \ \text{and}\ g(C_{0},I_{0},C_{1},I_{1})\\
        0              & \text{otherwise}
    \end{cases}
\end{equation}


\section{Experimental Setup}

We evaluate various configurations of the following multimodal transformers: CLIP \cite{radford2021clip}, FLAVA \cite{singh2022flava}, LXMERT \cite{tan2020lxmert}, UniT \cite{hu2021unit}, UNITER \cite{chen2020uniter}, VILLA \cite{gan2020villa}, VinVL \cite{zhang2021vinvl}, ViLT \cite{kim2021vilt}, VisualBERT \cite{li2019visualbert} and ViLBERT \cite{lu2019vilbert}. We also evaluate several configurations of two types of RNN-based models: VSE++ \cite{faghri2018vse} and VSRN \cite{li2019vsrn}.
We detail differences between these models and provide a high-level overview in \cref{tab:model-types}.
We also establish a human baseline using crowdworkers, as described in \cref{sec:human-perf}.

\subsection{Vision \& Language Transformers}\label{sec:vl-transformers}
\begin{table*}[!ht]
    \centering
    \small
    \resizebox{\textwidth}{!}{%
    \begin{tabular}{l|lr|l|l}
    \toprule
    \textit{Model} & \textit{Datasets} & \textit{\# Images, Captions (Millions)} & \textit{Architecture} & \textit{Attention} \\\midrule
    VinVL \cite{zhang2021vinvl}  & VQA, GQA, VG-QA, COCO, Flickr30k, CC, SBU & 1.89, 4.87 & single-stream & merged \\
    UNITER \cite{chen2020uniter}  & COCO, VG, CC, SBU & 4.20, 9.58 & single-stream  &  merged \\
    ViLLA \cite{gan2020villa} & COCO, VG, CC, SBU  & 4.20, 9.58 & single-stream  &  merged \\
    VisualBERT \cite{li2019visualbert}& COCO, NVLR2 & 0.30, 0.52  & single-stream  & merged \\
    ViLT \cite{kim2021vilt}  & COCO, VG, SBU, CC & 4.10, 9.85 & single-stream  & merged \\
    LXMERT \cite{tan2020lxmert}  & COCO, VG & 0.18, 9.18 & dual-stream & modality-specific, co-attn, merged \\
    ViLBERT \cite{lu2019vilbert}  & CC & 3.30, 3.30 & dual-stream  & modality-specific, co-attn, merged \\
    UniT \cite{hu2021unit} & COCO detect., VG detect., VQAv2, SNLI-VE QNLI, MNLI-mm, QQP, SST-2 & 0.69, 1.91 & dual-stream & modality-specific, merged\\
    FLAVA $_{ITM}$ \cite{singh2022flava}  & COCO, SBU, LN, CC, VG, WIT, CC 12M, RC, YFCC100M & 70.00, 70.00 & dual-stream & modality-specific, merged \\
    FLAVA $_{Contrastive}$ \cite{singh2022flava}  & COCO, SBU, LN, CC, VG, WIT, CC 12M, RC, YFCC100M & 70.00, 70.00 & dual-stream & modality-specific \\
    CLIP \cite{radford2021clip}  & $-$ & 400.00, 400.00 & dual-stream & modality-specific \\
    VSE++ and VSRN $_{COCO}$ & COCO & 0.11, 0.57 & dual-stream & $-$\\
    VSE++ and VSRN $_{Flickr30k}$ & Flickr30k & 0.03, 0.16 & dual-stream & $-$\\
    \bottomrule
    \end{tabular}
    }
    \caption{A high-level overview of the differences between the models we evaluate by the pretraining datasets, architecture, and attention mechanisms between the modalities. We omit datasets that were only used to train backbones. We exclude the language embedding from this table as every model uses a pretrained BERT tokenizer, except CLIP, VSE++, and VSRN. The pretraining datasets include COCO \cite{lin2014microsoft}, Visual Genome (VG) \cite{krishna2016visual}, Conceptual Captions (CC) \cite{sharma2018conceptual}, SBU Captions \cite{ordonez2011im2text}, Flickr30k \cite{young2014image}, VQA 2.0 \cite{goyal2017making}, VCR \cite{zellers2019recognition}, NLVR2 \cite{suhr2017corpus}, SNLI-VE \cite{xie2018visual}, QNLI \cite{rajpurkar2016squad}, MLNI-mm \cite{williams2017broad}, QQP \cite{QQPDataset}, Localized Narratives (LN) \cite{pont-tuset2020localized-narratives}, Wikipedia Image Text (WIT) \cite{srinivasan2021wit}, Conceptual Captions 12M (CC 12M) \cite{changpinyo2021conceptual12m}, Red Caps (RC) \cite{desai2021redcaps}, YFCC100M \cite{thomee2016yfcc100m}, and SST-2 \cite{Socher2013RecursiveDM}. CLIP uses their own dataset for pretraining.}
    \label{tab:model-types}
\end{table*}

\textbf{Image and language embedding.} All transformer models we evaluate use a pretrained BERT tokenizer \cite{devlin2019bert}, except CLIP, which uses a Byte-Pair Encoding tokenizer \cite{sennrich2015neural} trained from scratch. For the image embedding, five transformers (VisualBERT, ViLBERT, LXMERT, UNITER, ViLLA)\cite{li2019visualbert,lu2019vilbert,tan2020lxmert,chen2020uniter,gan2020villa} use region features extracted from the \texttt{fc6} layer of a Faster R-CNN~\cite{ren2015faster} trained on Visual Genome \cite{krishna2016visual}.
%
%
%
VinVL trains its own feature extractor on a large combined dataset from public sources with a unified object vocabulary \cite{zhang2021vinvl}.
The CLIP, FLAVA, and ViLT that we test all use Vision Transformer (ViT) \cite{dosovitskiy2020imageworth}.
In ViT, images are flattened into patches that are linearly projected and combined with a position encoding. UniT \cite{hu2021unit} alternatively uses a transformer network \cite{vaswani2017attention} on top of a convolutional network following Carion et al.\cite{carion2020end}.

\textbf{Single-stream vs. dual-stream encoders.} Vision and language transformers are mainly single- or dual-stream models: the embeddings for the image and text modalities are either concatenated and then jointly encoded (single-stream), or encoded by two separate modality-specific encoders with optional cross-modality fusion (dual-stream).
Five of our transformers are single-stream \cite{chen2020uniter, gan2020villa, zhang2021vinvl, kim2021vilt, li2019visualbert}. 
VinVL additionally concatenates object tags, which are the set of objects detected by the X152-C4 model during feature extraction, to the language tokens before encoding.
All single-stream models use merged attention, where the language and visual input attend to both themselves and the other modality.
The dual-stream transformers we evaluate are CLIP, FLAVA, UniT, LXMERT and ViLBERT \cite{radford2021clip, singh2022flava, hu2021unit, tan2020lxmert, lu2019vilbert}.
CLIP and the contrastive configuration of FLAVA lack cross-modal attention. 
ViLBERT has language-only transformer layers that are then fused by cross-modal transformer layers.
LXMERT, the ITM configuration of FLAVA, and UniT each use language-only and vision-only layers that are also fused by cross-modal transformer layers, which perform a combo of modality-specific attention and co-attention across modalities.
%

\textbf{Pretraining objectives.} V\&L transformers use a number of pretraining objectives including but not limited to masked language modeling, masked region modeling (classification of object classes and regression over image features) and image-text matching.
As we are evaluating a model's ability to determine if an image and a corresponding caption match, we select V\&L transformers that are pretrained with an image-text matching classification head or that produce a similarity score between the two modalities\footnote{UniT is the only model we selected that was not pretrained on image-text matching. To get image-text alignment scores, we finetuned UniT on image-text matching loss using MS-COCO\cite{lin2014microsoft}}.

\subsection{Multimodal RNNs}

To determine whether low performance on Winoground is unique to transformer-based models, we include results for two sequence-based models, which are VSRN\cite{li2019vsrn} and VSE++\cite{faghri2018vse}. Both VSE++ and VSRN have a loss function that prioritizes minimizing the hardest negative's score. The hardest negative is the highest-scoring image-caption pair that is not correct. Intuitively, this type of loss function could enable models to get higher scores on Winoground in particular and may be useful in future work. Although we show later in the paper that VSRN and VSE++ do not do well, perhaps due to issues besides the loss function. Both models use a GRU\cite{chung2014gru} to get language embeddings and a separate pipeline to get image embeddings. Scores for image-caption pairs are found by taking an inner-product of the embeddings. VSE's image encoder is a linear projection of the embedding from a backbone (either ResNet152\cite{he2016deep} or VGG19\cite{simonyan2015very}). In VSRN, a ResNet101-based Faster R-CNN with graph convolutions on top is used to get a sequence of features which are fed into a GRU. The GRU's last hidden state is then used as the image embedding.

\subsection{Human Performance}\label{sec:human-perf}

We employed crowd workers on the Amazon Mechanical Turk platform to establish a more conservative human baseline than the expert annotator upper bound of a perfect score.
Like the models, annotators are shown one image and one caption at a time. Annotators are asked the binary choice question ``Does the caption match the image?".
All 1600 combinations of images and captions are labeled by at least ten annotators.
We compute the human image-caption score as the ratio of annotators who said the image/caption pair match over the total number of annotators for the pair. More details about the human labelling interface, onboarding criteria, and quality control are provided in Appendix~E.

\section{Results}\label{sec:results}

\subsection{Compared to humans}
As shown in \cref{tab:results-aggr}, the models struggle across the board on Winoground, often performing close to or below random chance. Comparatively, as expected, the human performance is high across the full range of linguistic and visual phenomena.
For the \textbf{text score}, we observe $\sim$50\% absolute difference between humans and the best performing models---UNITER, VILLA VinVL, ViLT, FLAVA, and CLIP---with the remaining models below chance.

\begin{table}[t]
\centering
\resizebox{\columnwidth}{!}{%
\begin{tabular}{lrrr}
\toprule
 Model      & Text          & Image         & Group         \\
\midrule
 MTurk Human                  & \textbf{89.50} & \textbf{88.50} & \textbf{85.50}\\
 Random Chance                  & 25.00 & 25.00 & 16.67 \\\midrule
 VinVL                        & \textbf{37.75} & 17.75          & 14.50          \\
 UNITER$_{large}$             & \textbf{38.00} & 14.00          & 10.50          \\
 UNITER$_{base}$              & \textbf{32.25} & 13.25          & 10.00          \\
 ViLLA$_{large}$              & \textbf{37.00} & 13.25          & 11.00          \\
 ViLLA$_{base}$               & \textbf{30.00} & 12.00          & 8.00           \\
 VisualBERT$_{base}$          & 15.50          & 2.50           & 1.50           \\
 ViLT (ViT-B/32)              & \textbf{34.75} & 14.00          & 9.25           \\
 LXMERT                       & 19.25          & 7.00           & 4.00           \\
 ViLBERT$_{base}$             & 23.75          & 7.25           & 4.75           \\
 UniT$_{ITM finetuned}$       & 19.50          & 6.25           & 4.00           \\
 FLAVA$_{ITM}$                & \textbf{32.25} & 20.50 & 14.25 \\
 FLAVA$_{Contrastive}$        & \textbf{25.25} & 13.50          & 9.00           \\
 CLIP (ViT-B/32)              & \textbf{30.75} & 10.50          & 8.00           \\
 VSE++$_{COCO}$ (ResNet)      & 22.75          & 8.00           & 4.00           \\
 VSE++$_{COCO}$ (VGG)         & 18.75          & 5.50           & 3.50           \\
 VSE++$_{Flickr30k}$ (ResNet) & 20.00          & 5.00           & 2.75           \\
 VSE++$_{Flickr30k}$ (VGG)    & 19.75          & 6.25           & 4.50           \\
 VSRN$_{COCO}$                & 17.50          & 7.00           & 3.75           \\
 VSRN$_{Flickr30k}$           & 20.00          & 5.00           & 3.50           \\
\bottomrule
\end{tabular}
}
\caption{Results on the Winoground dataset across the text, image and group score metrics. Results above random chance in \textbf{bold}.}
\label{tab:results-aggr}
\end{table}

The human performance is only slightly lower for the \textbf{image score}, whereas all models perform much worse. Even the highest performing model, FLAVA$_{ITM}$, has a $\sim$70\% performance gap compared to humans. This gap is not unique to our dataset: in prior work \cite{faghri2018vse} \cite{radford2021clip}, models also tend to perform significantly better on caption retrieval compared to image retrieval. More investigation is required to pinpoint the reasons: perhaps textual encoders are stronger, or the text modality has different biases.

Lastly, we consider the \textbf{group score}. For humans, it is not appreciably lower than their text and image scores. All of the models are below random chance here as well.
We report confidence intervals for these results in Appendix~A.

\begin{table*}[!ht]
    \centering
    \resizebox{!}{1.32in}{%
  \begin{tabular}{lrrr|rrr|rrr|rrr|rrr}
    \toprule
     &
      \multicolumn{3}{c}{\textit{Object}} &
      \multicolumn{3}{c}{\textit{Relation}} &
      \multicolumn{3}{c}{\textit{Both}} &
      \multicolumn{3}{c}{\textit{1 Main Pred}} &
      \multicolumn{3}{c}{\textit{2 Main Preds}}\\
    \textit{Model} & Text & Image & Group & Text & Image & Group & Text & Image & Group & Text & Image & Group & Text & Image & Group \\\midrule
 MTurk Human                  & \textbf{92.20} & \textbf{90.78} & \textbf{88.65} & \textbf{89.27} & \textbf{90.56} & \textbf{86.70} & \textbf{76.92} & \textbf{57.69} & \textbf{57.69} & \textbf{87.33} & \textbf{85.62} & \textbf{82.53} & \textbf{95.37} & \textbf{96.30} & \textbf{93.52} \\
 VinVL                        & \textbf{36.88} & 17.73          & 14.18          & \textbf{37.77} & 17.60          & 14.16          & \textbf{42.31} & 19.23          & \textbf{19.23} & \textbf{39.38} & 21.23          & \textbf{17.47} & \textbf{33.33} & 8.33           & 6.48           \\
 UNITER$_{large}$             & \textbf{39.01} & 12.77          & 9.93           & \textbf{36.05} & 14.16          & 9.87           & \textbf{50.00} & 19.23          & \textbf{19.23} & \textbf{40.07} & 16.44          & 13.36          & \textbf{32.41} & 7.41           & 2.78           \\
 UNITER$_{base}$              & \textbf{34.04} & 11.35          & 9.22           & \textbf{30.04} & 14.16          & 10.30          & \textbf{42.31} & 15.38          & 11.54          & \textbf{35.27} & 14.73          & 11.99          & 24.07          & 9.26           & 4.63           \\
 ViLLA$_{large}$              & \textbf{36.88} & 14.89          & 11.35          & \textbf{37.34} & 12.88          & 11.16          & \textbf{34.62} & 7.69           & 7.69           & \textbf{39.73} & 17.12          & 14.38          & \textbf{29.63} & 2.78           & 1.85           \\
 ViLLA$_{base}$               & \textbf{33.33} & 15.60          & 9.93           & \textbf{27.04} & 9.01           & 6.01           & \textbf{38.46} & 19.23          & 15.38          & \textbf{33.22} & 14.04          & 10.27          & 21.30          & 6.48           & 1.85           \\
 VisualBERT$_{base}$          & 19.15          & 2.13           & 0.71           & 12.88          & 2.15           & 1.72           & 19.23          & 7.69           & 3.85           & 16.44          & 2.74           & 1.71           & 12.96          & 1.85           & 0.93           \\
 ViLT (ViT-B/32)              & \textbf{31.91} & 15.60          & 9.22           & \textbf{36.91} & 11.59          & 8.15           & \textbf{30.77} & \textbf{26.92} & \textbf{19.23} & \textbf{35.27} & 17.12          & 11.64          & \textbf{33.33} & 5.56           & 2.78           \\
 LXMERT                       & 22.70          & 9.22           & 6.38           & 17.60          & 5.58           & 2.58           & 15.38          & 7.69           & 3.85           & 19.18          & 8.56           & 5.14           & 19.44          & 2.78           & 0.93           \\
 ViLBERT$_{base}$             & \textbf{29.08} & 10.64          & 7.09           & 19.31          & 3.00           & 1.72           & \textbf{34.62} & \textbf{26.92} & \textbf{19.23} & 23.97          & 8.90           & 5.82           & 23.15          & 2.78           & 1.85           \\
 UniT$_{ITM finetuned}$       & 17.73          & 5.67           & 2.13           & 18.03          & 4.72           & 3.43           & \textbf{42.31} & 23.08          & \textbf{19.23} & 21.58          & 6.85           & 4.11           & 13.89          & 4.63           & 3.70           \\
  FLAVA$_{ITM}$                & \textbf{31.91} & 23.40 & 14.89 & \textbf{30.04} & 16.31 & 12.02 & \textbf{53.85} & \textbf{42.31} & \textbf{30.77} & \textbf{36.30} & 24.66 & \textbf{17.81} & 21.30          & 9.26 & 4.63 \\
 FLAVA$_{Contrastive}$        & 23.40          & 19.15 & 11.35 & 23.61          &  8.58 &  5.58 & \textbf{50.00} & \textbf{26.92} & \textbf{26.92} & \textbf{26.37} & 16.44 & 10.62          & 22.22          &  5.56 & 4.63 \\
 CLIP (ViT-B/32)              & \textbf{34.75} & 7.80           & 6.38           & 22.75          & 8.58           & 5.58           & \textbf{80.77} & \textbf{42.31} & \textbf{38.46} & \textbf{35.27} & 13.01          & 10.27          & 18.52          & 3.70           & 1.85           \\
 VSE++$_{COCO}$ (ResNet)      & 21.99          & 6.38           & 1.42           & 23.61          & 9.01           & 5.58           & 19.23          & 7.69           & 3.85           & 25.00          & 9.59           & 4.79           & 16.67          & 3.70           & 1.85           \\
 VSE++$_{COCO}$ (VGG)         & 17.73          & 2.13           & 2.13           & 18.45          & 7.30           & 3.86           & \textbf{26.92} & 7.69           & 7.69           & 18.49          & 4.79           & 2.74           & 19.44          & 7.41           & 5.56           \\
 VSE++$_{Flickr30k}$ (ResNet) & 20.57          & 6.38           & 3.55           & 18.88          & 4.29           & 2.15           & \textbf{26.92} & 3.85           & 3.85           & 21.58          & 6.51           & 3.42           & 15.74          & 0.93           & 0.93           \\
 VSE++$_{Flickr30k}$ (VGG)    & 17.73          & 4.96           & 2.84           & 19.74          & 6.87           & 5.15           & \textbf{30.77} & 7.69           & 7.69           & 20.55          & 6.16           & 4.79           & 17.59          & 6.48           & 3.70           \\
 VSRN$_{COCO}$                & 15.60          & 4.96           & 2.13           & 18.88          & 7.73           & 4.72           & 15.38          & 11.54          & 3.85           & 17.12          & 7.19           & 3.77           & 18.52          & 6.48           & 3.70           \\
 VSRN$_{Flickr30k}$           & 16.31          & 4.96           & 2.13           & 21.03          & 4.29           & 3.86           & \textbf{30.77} & 11.54          & 7.69           & 20.89          & 5.82           & 3.77           & 17.59          & 2.78           & 2.78           \\
          \bottomrule
  \end{tabular}
  }
  \caption{The results by linguistic tag. Results above chance are in \textbf{bold}.}
    \label{tab:results-by-ling-tag}
\end{table*}

\begin{table*}
    \centering
    \resizebox{!}{1.32in}{%
  \begin{tabular}{lrrr|rrr|rrr}
    \toprule
     &
      \multicolumn{3}{c}{\textit{Symbolic}} &
      \multicolumn{3}{c}{\textit{Pragmatics}} &
      \multicolumn{3}{c}{\textit{Same Image Series}} \\
    \textit{Model} & Text & Image & Group & Text & Image & Group & Text & Image & Group \\\midrule
 MTurk Human                  & \textbf{96.43} & \textbf{92.86} & \textbf{92.86} & \textbf{58.82} & \textbf{41.18} & \textbf{41.18} & \textbf{95.65} & \textbf{91.30} & \textbf{91.30} \\
 VinVL                        & 25.00          & 17.86          & 14.29          & \textbf{29.41} & 5.88           & 5.88           & \textbf{34.78} & 17.39          & 13.04          \\
 UNITER$_{large}$             & \textbf{39.29} & \textbf{28.57} & \textbf{17.86} & \textbf{35.29} & 0.00           & 0.00           & 4.35           & 8.70           & 0.00           \\
 UNITER$_{base}$              & \textbf{46.43} & 14.29          & 14.29          & \textbf{29.41} & 17.65          & 11.76          & 8.70           & 8.70           & 0.00           \\
 ViLLA$_{large}$              & \textbf{39.29} & 14.29          & 10.71          & 17.65          & 0.00           & 0.00           & 17.39          & 4.35           & 0.00           \\
 ViLLA$_{base}$               & \textbf{42.86} & 17.86          & 14.29          & \textbf{29.41} & 5.88           & 5.88           & 13.04          & 8.70           & 4.35           \\
 VisualBERT$_{base}$          & \textbf{28.57} & 0.00           & 0.00           & 5.88           & 0.00           & 0.00           & 13.04          & 0.00           & 0.00           \\
 ViLT (ViT-B/32)              & \textbf{28.57} & 17.86          & 10.71          & \textbf{35.29} & 0.00           & 0.00           & \textbf{26.09} & 0.00           & 0.00           \\
 LXMERT                       & \textbf{28.57} & 3.57           & 3.57           & 17.65          & 5.88           & 0.00           & 8.70           & 4.35           & 0.00           \\
 ViLBERT$_{base}$             & \textbf{28.57} & 10.71          & 7.14           & \textbf{29.41} & 5.88           & 5.88           & 13.04          & 0.00           & 0.00           \\
 UniT$_{ITM finetuned}$       & 14.29          & 10.71          & 7.14           & 17.65          & 5.88           & 5.88           & 21.74          & 4.35           & 4.35           \\
 FLAVA$_{ITM}$                & 25.00          & \textbf{28.57} & \textbf{17.86} & 17.65          & \textbf{29.41} & 11.76 & 17.39          &  8.70 &  0.00 \\
 FLAVA$_{Contrastive}$        & 17.86          & 10.71          & 10.71          & 11.76          & 23.53          & 5.88           & 17.39          & 4.35           &  4.35 \\
 CLIP (ViT-B/32)              & \textbf{39.29} & 3.57           & 3.57           & \textbf{35.29} & 5.88           & 5.88           & 8.70           & 0.00           & 0.00           \\
 VSE++$_{COCO}$ (ResNet)      & \textbf{32.14} & 10.71          & 10.71          & 23.53          & 11.76          & 0.00           & 13.04          & 4.35           & 4.35           \\
 VSE++$_{COCO}$ (VGG)         & 17.86          & 14.29          & 7.14           & 17.65          & 0.00           & 0.00           & 13.04          & 4.35           & 4.35           \\
 VSE++$_{Flickr30k}$ (ResNet) & 21.43          & 3.57           & 0.00           & 23.53          & 0.00           & 0.00           & 17.39          & 4.35           & 0.00           \\
 VSE++$_{Flickr30k}$ (VGG)    & \textbf{28.57} & 10.71          & 10.71          & 11.76          & 0.00           & 0.00           & 13.04          & 4.35           & 0.00           \\
 VSRN$_{COCO}$                & 7.14           & 3.57           & 0.00           & 11.76          & 0.00           & 0.00           & 13.04          & 0.00           & 0.00           \\
 VSRN$_{Flickr30k}$           & 21.43          & 3.57           & 3.57           & \textbf{35.29} & 11.76          & 5.88           & 8.70           & 4.35           & 4.35           \\
    \bottomrule
  \end{tabular}
  }
  \caption{The results by visual tag. Results above chance are in \textbf{bold}.}
    \label{tab:results-by-visual-tag}
\end{table*}

\subsection{Results by Tags}

For the swap-dependent linguistic tags, human performance is highest on \textbf{object}, followed by the \textbf{relation} and then \textbf{both}. For the swap-independent linguistic tags, humans do better on examples with two main predicates, which tend to contain longer and more complicated sentences. The models perform poorly on every category, but they largely show the opposite pattern. They perform better on examples with simpler and shorter sentences which more often have swaps at the morpheme level (see \cref{tab:results-by-ling-tag}). One exception to the low model performance is that CLIP performs comparably to the humans on the \textbf{both} tag text score---the 26 examples with the \textbf{both} tag have some of the shortest and least compositional captions in our dataset (e.g. ``presenting the watch'' vs ``watching the present''). 

We also evaluate performance for the visual reasoning tags as shown in \cref{tab:results-by-visual-tag}. Models and humans are particularly good at the \textbf{symbolic} examples, but the models are poor comparatively. On the \textbf{pragmatics} tag, humans have the lowest performance. Ten crowdworkers probably didn't capture slight pragmatics preferences that our expert linguist annotators agreed on. One example that the crowdworkers failed is \cref{fig:dataset-examples}(a): “the kid [with the magnifying glass] looks at them []”. All ten annotators said that ``the kid with the magnifying glass looks at them" was acceptable for both images, but captured the correct preference for the second caption. This reveals a limitation in how the task was presented to humans: our hypothesis is that if we gave humans both images and both captions at the same time, or if significantly more human annotators gave their judgements, then the human scores would be substantially higher. Finally, models do worst on the \textbf{series} tag where most get a 0\% group score, which indicates that they are always choosing one image over the other regardless of the caption (or vice versa).

\begin{figure*}[t]
    \centering
    \begin{minipage}{0.48\textwidth}
        \centering
        \scalebox{0.9}{
            \begin{tabular}{cccc}
            \multicolumn{3}{c}{a \textbf{brown dog} is on a \textbf{white couch}}\\
             \includegraphics[height=1.5cm, width=0.25\textwidth]{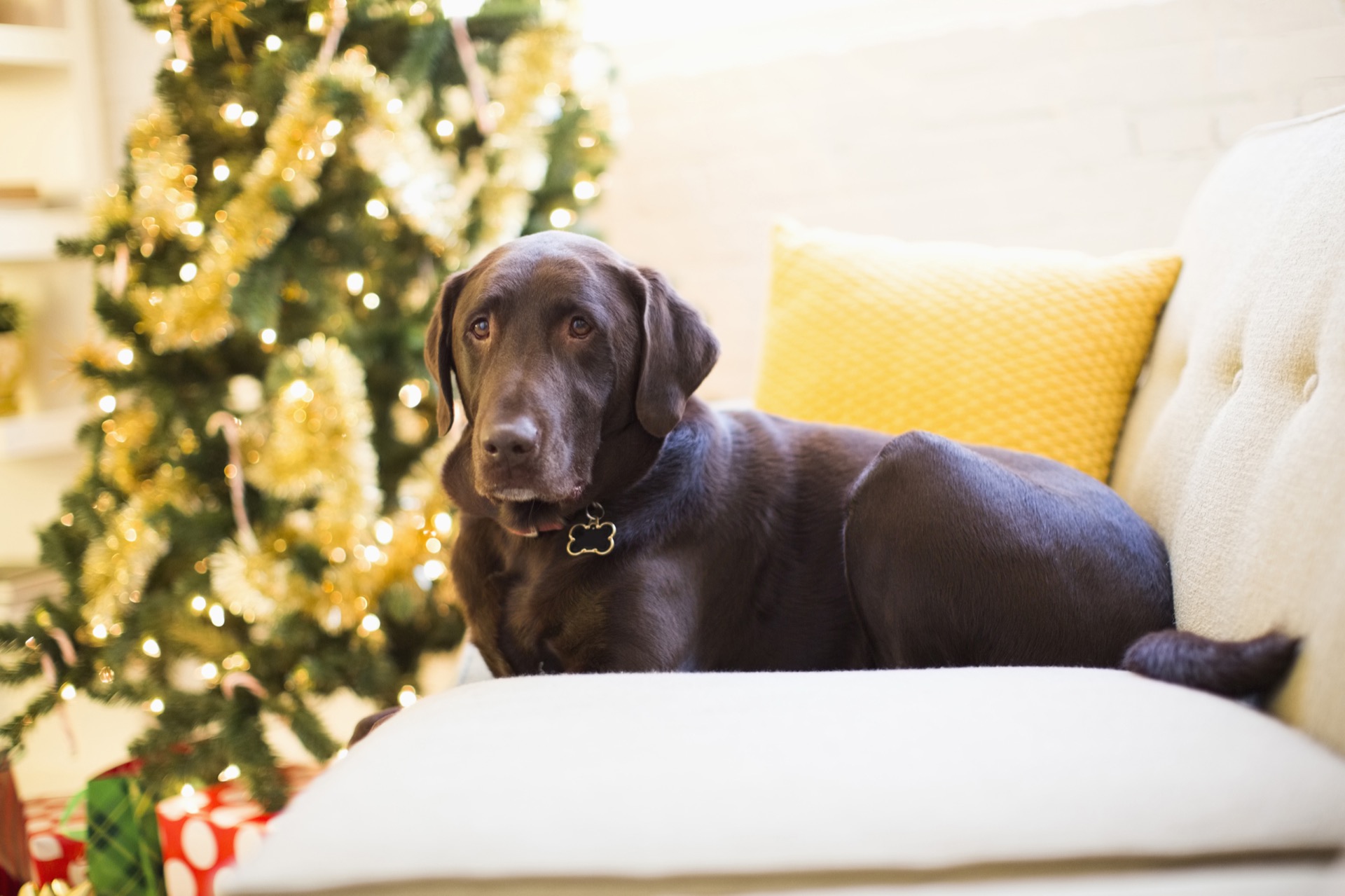}& \includegraphics[height=1.5cm, width=0.25\textwidth]{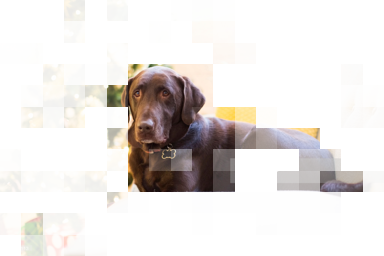} & \includegraphics[height=1.5cm, width=0.25\textwidth]{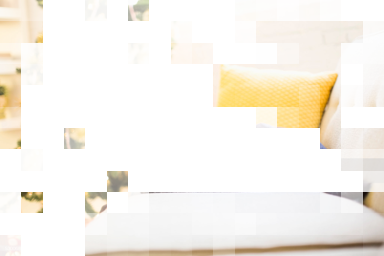} \\
             \includegraphics[height=1.5cm, width=0.25\textwidth]{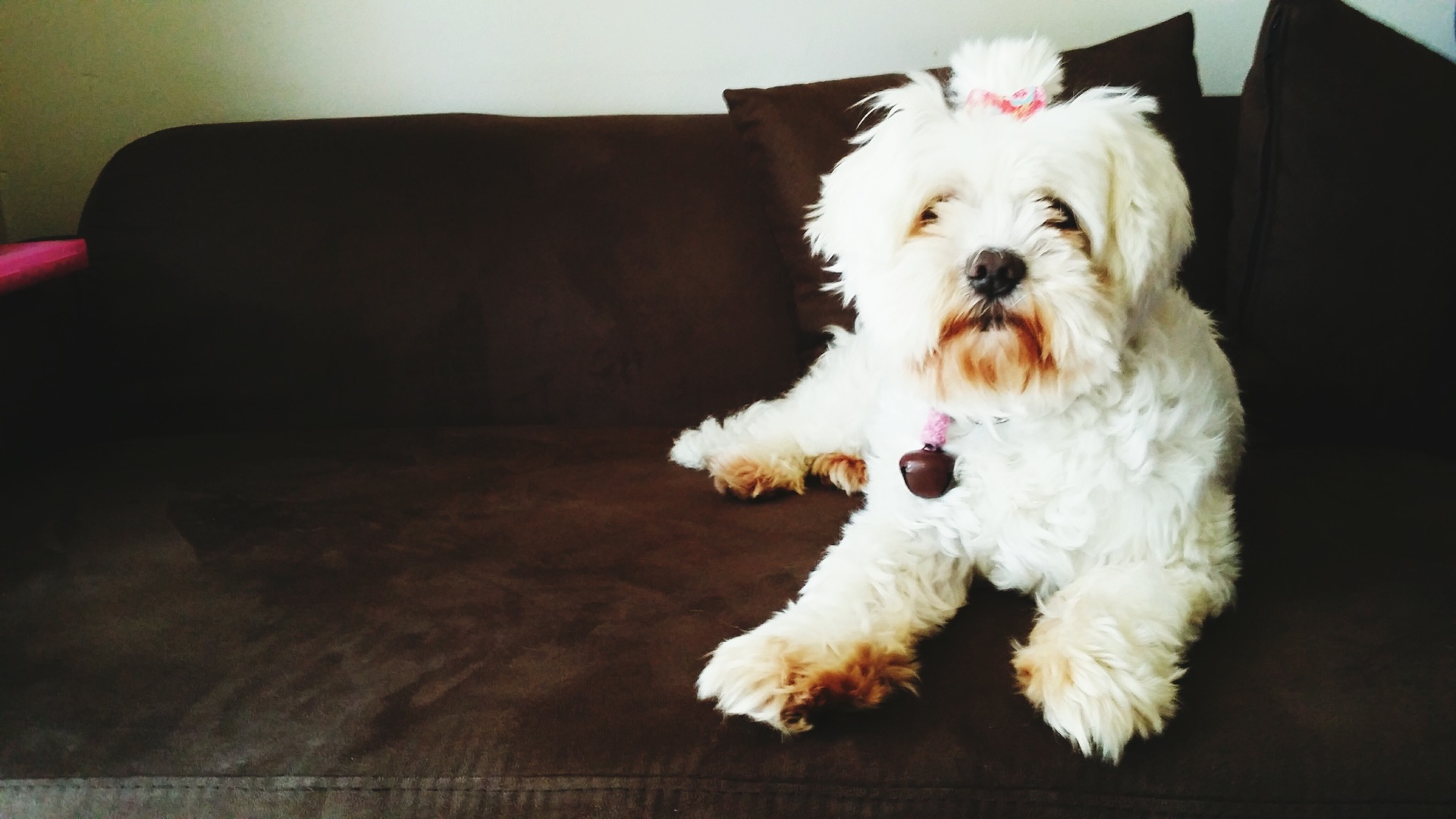}& \includegraphics[height=1.5cm, width=0.25\textwidth]{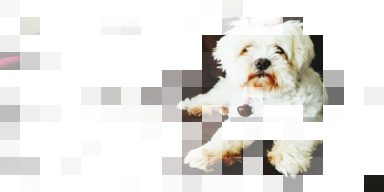} & \includegraphics[height=1.5cm, width=0.25\textwidth]{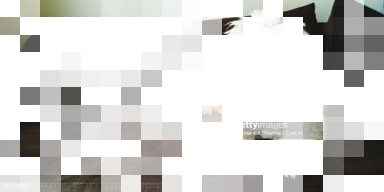} \\
             & & & \\
             \multicolumn{3}{c}{a \textbf{white dog}\quad is on a \quad\textbf{brown couch}}\\
             \includegraphics[height=1.5cm, width=0.25\textwidth]{vilt_ot_heatmap/e29_i1_base.png}& \includegraphics[height=1.5cm, width=0.25\textwidth]{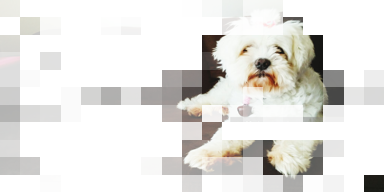} & \includegraphics[height=1.5cm, width=0.25\textwidth]{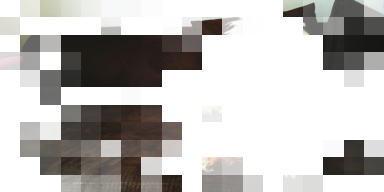} \\
             \includegraphics[height=1.5cm, width=0.25\textwidth]{vilt_ot_heatmap/e29_i0_base.png}& \includegraphics[height=1.5cm, width=0.25\textwidth]{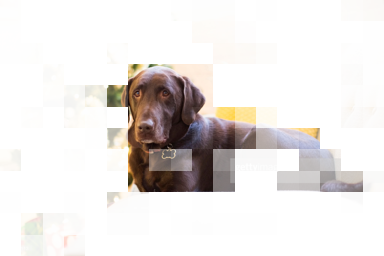} & \includegraphics[height=1.5cm, width=0.25\textwidth]{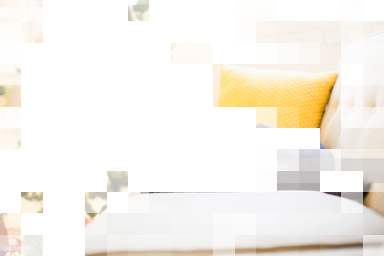} \\
            \end{tabular}
        }
    \end{minipage}
    ~
    \begin{minipage}{0.48\textwidth}
        \centering
        \scalebox{0.9}{
            \begin{tabular}{cccc}
            \multicolumn{3}{c}{\textbf{circular food} on \textbf{heart-shaped wood}}\\
            \includegraphics[height=1.5cm, width=0.20\textwidth]{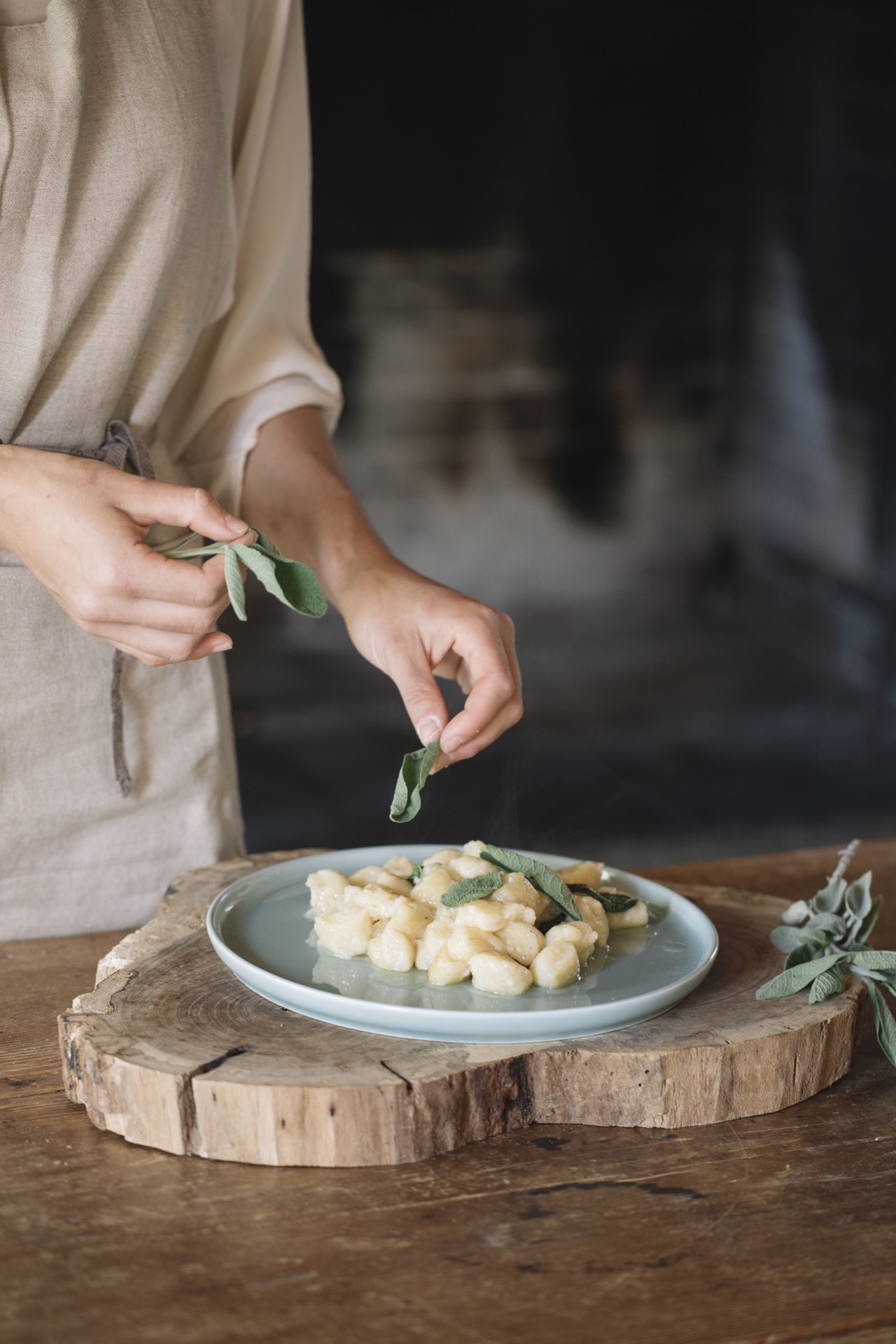}& \includegraphics[height=1.5cm, width=0.20\textwidth]{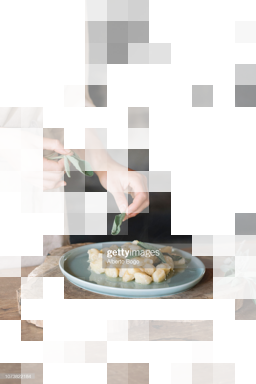} & \includegraphics[height=1.5cm, width=0.20\textwidth]{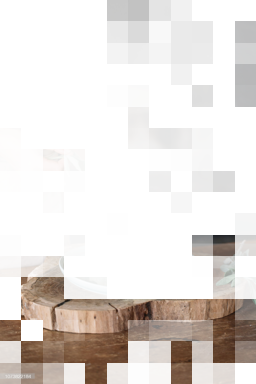} \\
            \includegraphics[height=1.5cm, width=0.25\textwidth]{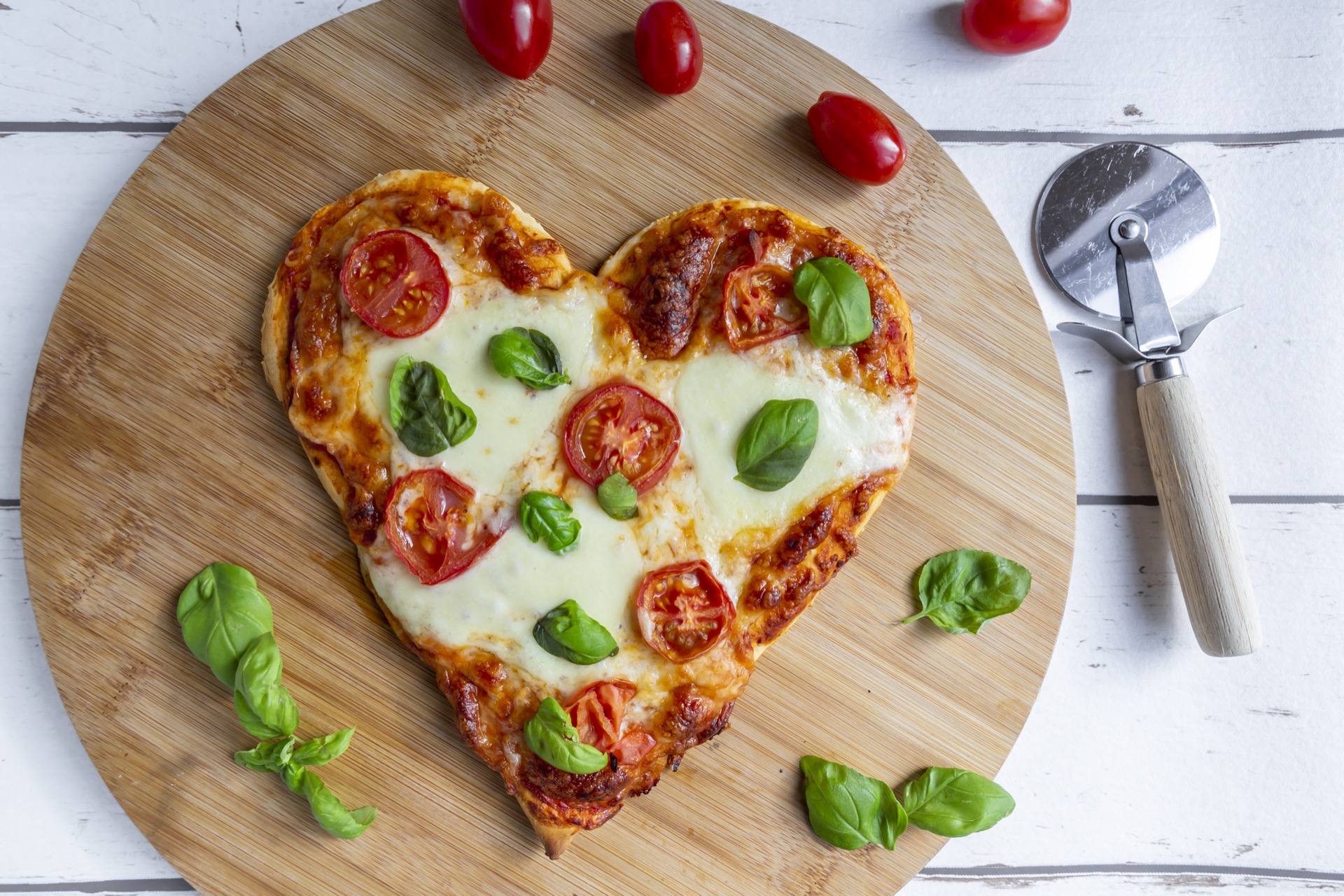}& \includegraphics[height=1.5cm, width=0.25\textwidth]{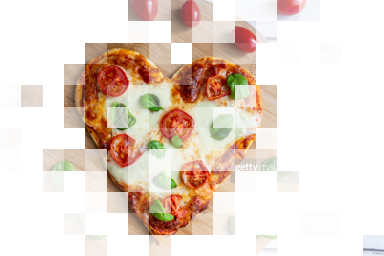} & \includegraphics[height=1.5cm, width=0.25\textwidth]{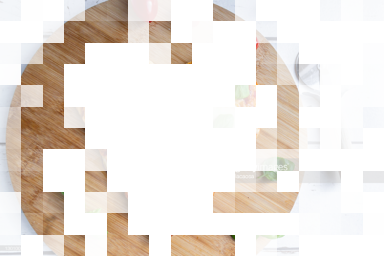} \\
             & & & \\
             \multicolumn{3}{c}{ \textbf{heart-shaped food} on \textbf{circular wood}}\\
            \includegraphics[height=1.5cm, width=0.25\textwidth]{vilt_ot_heatmap/e118_img_1.png}& \includegraphics[height=1.5cm, width=0.25\textwidth]{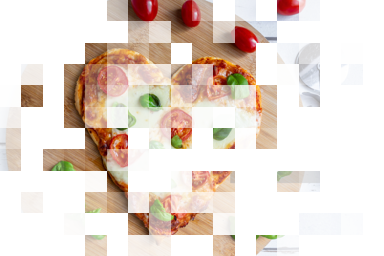} & \includegraphics[height=1.5cm, width=0.25\textwidth]{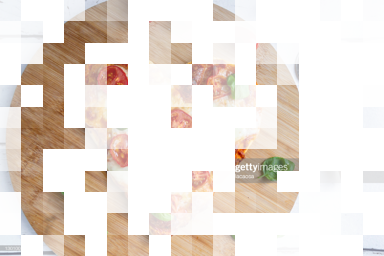}.png \\
            \includegraphics[height=1.5cm, width=0.20\textwidth]{vilt_ot_heatmap/e118_img_0.png}& \includegraphics[height=1.5cm, width=0.20\textwidth]{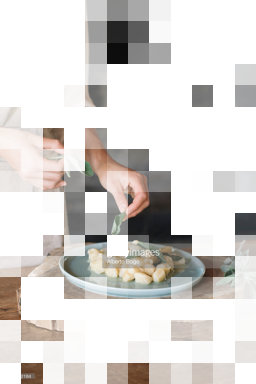} & \includegraphics[height=1.5cm, width=0.20\textwidth]{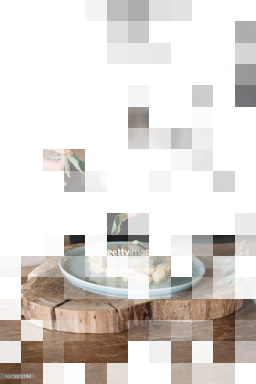} \\
            \end{tabular}
        }
    \end{minipage}
    
    \caption{Word-region alignment scores between the image and text features for ViLT \cite{kim2021vilt} on examples from Winoground. In this case study, ViLT appears to disregard the information from adjectives. E.g., the heatmaps highlight the brown dog just as strongly regardless of whether the text was ``brown dog" or ``white dog".}
    \label{fig:vilt-heatmap}
\end{figure*}

\begin{table}[t]
\centering
\resizebox{\columnwidth}{!}{%
\begin{tabular}{lrrrr}
\toprule
 & \multicolumn{2}{c}{\textit{Perplexity}} &  \multicolumn{2}{c}{\textit{Caption Length}}\\
 \textit{Model}      &   Corr. &   p-value & Corr. &   p-value\\\midrule 
 MTurk Human                  &  0.05 & 0.07 &  \textbf{0.20} & \textbf{0.00} \\
 VinVL                        & \textbf{-0.05} & \textbf{0.04} & \textbf{-0.20} & \textbf{0.00} \\
 UNITER$_{large}$             & -0.01 & 0.57 & \textbf{-0.16} & \textbf{0.00} \\
 UNITER$_{base}$              & -0.03 & 0.22 & \textbf{-0.14} & \textbf{0.00} \\
 ViLLA$_{large}$              & -0.02 & 0.39 & \textbf{-0.12} & \textbf{0.01} \\
 ViLLA$_{base}$               & -0.04 & 0.13 & \textbf{-0.11} & \textbf{0.03} \\
 VisualBERT$_{base}$          & -0.04 & 0.15 & -0.06 & 0.22 \\
 ViLT (ViT-B/32)              & -0.04 & 0.16 & \textbf{-0.16} & \textbf{0.00} \\
 LXMERT                       & -0.04 & 0.12 & \textbf{-0.11} & \textbf{0.02} \\
 ViLBERT$_{base}$             & -0.04 & 0.11 & \textbf{-0.14} & \textbf{0.00} \\
 UniT$_{ITM finetuned}$       & -0.01 & 0.73 & -0.02 & 0.73 \\
 FLAVA$_{ITM}$                & -0.03 & 0.22 & \textbf{-0.23} & \textbf{0.00} \\
 FLAVA$_{Contrastive}$        & \textbf{-0.06} & \textbf{0.01} & \textbf{-0.19} & \textbf{0.00} \\
 CLIP (ViT-B/32)              & -0.04 & 0.09 & \textbf{-0.22} & \textbf{0.00} \\
 VSE++$_{COCO}$ (ResNet)      & \textbf{-0.05} & \textbf{0.04} &  0.01 & 0.90 \\
 VSE++$_{COCO}$ (VGG)         & -0.04 & 0.08 &  0.03 & 0.56 \\
 VSE++$_{Flickr30k}$ (ResNet) & -0.02 & 0.43 &  0.02 & 0.67 \\
 VSE++$_{Flickr30k}$ (VGG)    &  0.01 & 0.74 & \textbf{-0.10} & \textbf{0.04} \\
 VSRN$_{COCO}$                & \textbf{-0.07} & \textbf{0.01} & -0.05 & 0.36 \\
 VSRN$_{Flickr30k}$           & -0.02 & 0.32 & -0.05 & 0.29 \\
\bottomrule
\end{tabular}
}
\caption{(left) The correlation between model image-caption scores and the caption perplexity from GPT2. (right) The correlation between the model group scores and the caption length.}
\label{tab:perplexity-and-length-correlations}
\end{table}

\begin{table}[t]
\centering
\begin{tabular}{llrr}
\toprule
Pretraining Modality & Score & Corr. & p-value\\\midrule
 & Text & \textbf{0.84} & \textbf{0.00} \\
 Image & Image & \textbf{0.76} & \textbf{0.00} \\
 & Group & \textbf{0.75} & \textbf{0.00} \\\midrule
 & Text  & \textbf{0.77} & \textbf{0.00} \\
 Caption & Image & \textbf{0.75} & \textbf{0.00} \\
 & Group & \textbf{0.71} & \textbf{0.00} \\
\bottomrule
\end{tabular}
\caption{Correlations between the number of pretraining images and captions and the model text, image, and group scores. CLIP and FLAVA are excluded as outliers.}
\label{tab:data-size-correlations}
\end{table}

\section{Discussion}

Despite the fact that every model struggled on Winoground compared to humans, we hope to gain further insights by analyzing which aspects of these models could contribute to their performance differences.

\subsection{Capabilities of Encoders}

\textbf{Richer features.} UNITER, VILLA, VinVL, ViLT, FLAVA, and CLIP are the only models that get above random chance performance in \cref{tab:results-aggr}, and only for the text score.
We hypothesize that these models perform better than others due to their richer features (unimodal features for CLIP and FLAVA$_{Contrastive}$, multimodal features for the others).
%
A potential explanation could be the large-scale pretraining used by CLIP and FLAVA, the large training dataset used to train the object detector for VinVL, or the ViT approach for image features used by ViLT, FLAVA, and CLIP that encodes every portion of the image.
%

\textbf{Common failure modes.} We highlight again that most of the models fail with 0\% group score on the \textit{same image series} tag.
One explanation is that the models' visual encoders might be too weak to correctly discriminate between substantially similar images. This could cause the models to fall back on their unimodal priors, picking one caption or image over the other in the majority of the four potential caption-image pairings.
%

\textbf{Heat maps.} We show a heatmap in \cref{fig:vilt-heatmap} of the word-region alignment between ViLT's vision and language features as a visualization for a model with some of the better performance on our dataset. ViLLA and UNITER are also trained with word-region alignment and we provide their heatmaps in Appendix~D.

%

\textbf{Complicated captions.} The above-chance models do worse on examples with longer captions, possibly due to weak language encoding abilities. As shown in \cref{tab:perplexity-and-length-correlations}, caption length and lower model performance significantly correlate for the best models, even though the correlation is reversed for humans. The examples with the shortest captions are also the least compositional; they are primarily the examples where the parts of speech change between swapped words, or where there is a morpheme-level swap. Finally, we show in \cref{tab:perplexity-and-length-correlations} correlations between caption perplexity\footnote{We used the standard size GPT2 checkpoint from Hugging Face transformers to get perplexity\cite{wolf-etal-2020-transformers}.} and model scores. We found that there is typically a weak correlation between models assigning an image-caption pair a higher score and a caption having low perplexity.

\subsection{By Architecture \& Type of Attention}

%
As shown in \cref{tab:results-aggr,tab:results-by-ling-tag,tab:results-by-visual-tag}, both single-stream and dual-stream models perform significantly worse than humans on the text, image and group scores.
We find at least one single-stream model and at least one dual-stream model are above chance for most of our experiments, suggesting there is not a distinct performance difference by architecture. Although, six single-stream model checkpoints do above chance overall, compared to only the very large dual-stream models (CLIP and FLAVA). CLIP and FLAVA were trained on an order of magnitude more data than the other models.
Across all types of attention, models struggled compared to humans.
%
%
%
But neither of the two models using co-attention, in conjunction with single-modality and/or merged attention, performed above chance.
%
%
%

\subsection{By Multimodal Pretraining Dataset Size}

We find highly significant correlations between the size of the multimodal pretraining dataset and the scores, if we remove CLIP and FLAVA as outliers. \cref{tab:data-size-correlations} shows these correlations, and Appendix~B has graphs showing each model's score versus the pretraining data size. The unimodal training data (for image backbones or pre-initialized text encoders) is not included in these calculations.



\section{Conclusion}

We introduced a novel task and dataset, Winoground, aimed at measuring visio-linguistic compositional reasoning in state of the art vision and language models. We demonstrate that models fall short, in most cases performing no better than chance. Our findings highlight that there is more work to be done. Particularly, the field could investigate possible strengths of single-stream models, the compilation of more pretraining data, improving image-encoding capabilities, and pretraining objectives that emphasize similar but wrong images. We hope that our task and dataset will help guide research in this important direction.

\textbf{Broader Impact \& Limitations.} Winoground is English-only and translation to other languages may be nontrivial \cite{liu-etal-2021-visually}.
Expert curation is time-consuming and our dataset is limited in size.
Multimodal datasets containing images of people require thoughtful consideration of how people are represented (see \cite{birhane2021multimodal} for a detailed analysis of the stereotypes present in many multimodal datasets).
We used gender underspecified human denoting terms (e.g., person, child) to avoid issues with inferring gender identity from images \cite{Scheuerman2019gender}.
Our annotators disproportionately come from the USA and the same could be true for our crowdworkers.

\textbf{Getty Acknowledgement.} Images in the paper are a compilation of assets, including \copyright Getty Images/Natasha Breen, Maki Nakamura, Jessica Peterson, Kundanlall Sharma, lacaosa, Alberto Bogo, Vu Le, Toson Rueangsuksut, Nisian Hughes, Tanja Walter, Douglas Sacha, PBNJ Productions, Glow Images, 10'000 Hours, zoranm, Marlene Ford, Westend61.




{\small
\bibliographystyle{ieee_fullname}
\bibliography{wino}

\begin{thebibliography}{10}\itemsep=-1pt

\bibitem{akula2020words}
Arjun Akula, Spandana Gella, Yaser Al-Onaizan, Song-Chun Zhu, and Siva Reddy.
\newblock Words aren’t enough, their order matters: On the robustness of
  grounding visual referring expressions.
\newblock In {\em ACL}, 2020.

\bibitem{altshuler2019course}
Daniel Altshuler, Terence Parsons, and Roger Schwarzschild.
\newblock {\em A Course in Semantics}.
\newblock MIT Press, 2019.

\bibitem{antol2015vqa}
Stanislaw Antol, Aishwarya Agrawal, Jiasen Lu, Margaret Mitchell, Dhruv Batra,
  C~Lawrence Zitnick, and Devi Parikh.
\newblock Vqa: Visual question answering.
\newblock In {\em ICCV}, 2015.

\bibitem{bender2015establishing}
David Bender.
\newblock Establishing a human baseline for the winograd schema challenge.
\newblock In {\em Modern Artificial Intelligence and Cognitive Science}, 2015.

\bibitem{birhane2021multimodal}
Abeba Birhane, Vinay~Uday Prabhu, and Emmanuel Kahembwe.
\newblock Multimodal datasets: misogyny, pornography, and malignant
  stereotypes.
\newblock In {\em arXiv preprint arXiv:2110.01963}, 2021.

\bibitem{bitton-etal-2021-automatic}
Yonatan Bitton, Gabriel Stanovsky, Roy Schwartz, and Michael Elhadad.
\newblock Automatic generation of contrast sets from scene graphs: Probing the
  compositional consistency of {GQA}.
\newblock In {\em NAACL: Human Language Technologies}, 2021.

\bibitem{bogin2021covr}
Ben Bogin, Shivanshu Gupta, Matt Gardner, and Jonathan Berant.
\newblock Covr: A test-bed for visually grounded compositional generalization
  with real images.
\newblock In {\em EMNLP}, 2021.

\bibitem{cao2020behind}
Jize Cao, Zhe Gan, Yu Cheng, Licheng Yu, Yen-Chun Chen, and Jingjing Liu.
\newblock Behind the scene: Revealing the secrets of pre-trained
  vision-and-language models.
\newblock In {\em ECCV}, 2020.

\bibitem{carion2020end}
Nicolas Carion, Francisco Massa, Gabriel Synnaeve, Nicolas Usunier, Alexander
  Kirillov, and Sergey Zagoruyko.
\newblock End-to-end object detection with transformers.
\newblock In {\em ECCV}, 2020.

\bibitem{changpinyo2021conceptual12m}
Soravit Changpinyo, Piyush Sharma, Nan Ding, and Radu Soricut.
\newblock Conceptual 12m: Pushing web-scale image-text pre-training to
  recognize long-tail visual concepts.
\newblock In {\em CVPR}, 2021.

\bibitem{chao2017being}
Wei-Lun Chao, Hexiang Hu, and Fei Sha.
\newblock Being negative but constructively: Lessons learnt from creating
  better visual question answering datasets.
\newblock In {\em arXiv preprint arXiv:1704.07121}, 2017.

\bibitem{chen2020uniter}
Yen-Chun Chen, Linjie Li, Licheng Yu, Ahmed~El Kholy, Faisal Ahmed, Zhe Gan, Yu
  Cheng, and Jingjing Liu.
\newblock Uniter: Universal image-text representation learning.
\newblock In {\em ECCV}, 2020.

\bibitem{choi2012context}
Myung~Jin Choi, Antonio Torralba, and Alan~S. Willsky.
\newblock Context models and out-of-context objects.
\newblock In {\em Pattern Recognition Letters}, 2012.

\bibitem{chung2014gru}
Junyoung Chung, Caglar Gulcehr, KyungHyun Cho, and Yoshua Bengio.
\newblock Empirical evaluation of gated recurrent neural networks on sequence
  modeling.
\newblock In {\em NeurIPS}, 2014.

\bibitem{desai2021redcaps}
Karan Desai, Gaurav Kaul, Zubin Aysola, and Justin Johnson.
\newblock Redcaps: Web-curated image-text data created by the people.
\newblock In {\em NeurIPS Datasets and Benchmarks}, 2021.

\bibitem{devlin2019bert}
Jacob Devlin, Ming-Wei Chang, Kenton Lee, and Kristina Toutanova.
\newblock {BERT}: Pre-training of deep bidirectional transformers for language
  understanding.
\newblock In {\em NAACL: Human Language Technologies}, 2019.

\bibitem{ding2016understanding}
Nan Ding, Sebastian Goodman, Fei Sha, and Radu Soricut.
\newblock Understanding image and text simultaneously: a dual vision-language
  machine comprehension task.
\newblock In {\em arXiv preprint arXiv:1612.07833}, 2016.

\bibitem{dosovitskiy2020imageworth}
Alexey Dosovitskiy, Lucas Beyer, Alexander Kolesnikov, Dirk Weissenborn,
  Xiaohua Zhai, Thomas Unterthiner, Mostafa Dehghani, Matthias Minderer, Georg
  Heigold, Sylvain Gelly, Jakob Uszkoreit, and Neil Houlsby.
\newblock {An Image is Worth 16x16 Words: Transformers for Image Recognition at
  Scale}.
\newblock In {\em ICLR}, 2021.

\bibitem{dou2021empirical}
Zi-Yi Dou, Yichong Xu, Zhe Gan, Jianfeng Wang, Shuohang Wang, Lijuan Wang,
  Chenguang Zhu, Zicheng Liu, Michael Zeng, et~al.
\newblock An empirical study of training end-to-end vision-and-language
  transformers.
\newblock In {\em arXiv preprint arXiv:2111.02387}, 2021.

\bibitem{elazar2021back}
Yanai Elazar, Hongming Zhang, Yoav Goldberg, and Dan Roth.
\newblock Back to square one: Artifact detection, training and commonsense
  disentanglement in the winograd schema.
\newblock In {\em EMNLP}, 2021.

\bibitem{faghri2018vse}
Fartash Faghri, David~J. Fleet, Jamie~Ryan Kiros, and Sanja Fidler.
\newblock Vse++: Improving visual-semantic embeddings with hard negatives.
\newblock In {\em BMVC}, 2018.

\bibitem{frank-etal-2021-vision}
Stella Frank, Emanuele Bugliarello, and Desmond Elliott.
\newblock Vision-and-language or vision-for-language? on cross-modal influence
  in multimodal transformers.
\newblock In {\em EMNLP}, 2021.

\bibitem{gan2020villa}
Zhe Gan, Yen-Chun Chen, Linjie Li, Chen Zhu, Yu Cheng, and Jingjing Liu.
\newblock Large-scale adversarial training for vision-and-language
  representation learning.
\newblock In {\em NeurIPS}, 2020.

\bibitem{gauthier-etal-2020-syntaxgym}
Jon Gauthier, Jennifer Hu, Ethan Wilcox, Peng Qian, and Roger Levy.
\newblock {S}yntax{G}ym: An online platform for targeted evaluation of language
  models.
\newblock In {\em ACL: System Demonstrations}, 2020.

\bibitem{getty}
\url{https://www.gettyimages.com/}.

\bibitem{gomez2020exploring}
Raul Gomez, Jaume Gibert, Lluis Gomez, and Dimosthenis Karatzas.
\newblock Exploring hate speech detection in multimodal publications.
\newblock In {\em ICCV}, 2020.

\bibitem{goyal2017making}
Yash Goyal, Tejas Khot, Douglas Summers-Stay, Dhruv Batra, and Devi Parikh.
\newblock Making the v in vqa matter: Elevating the role of image understanding
  in visual question answering.
\newblock In {\em CVPR}, 2017.

\bibitem{gulordava2018}
Kristina Gulordava, Piotr Bojanowski, Edouard Grave, Tal Linzen, and Marco
  Baroni.
\newblock Colorless green recurrent networks dream hierarchically.
\newblock In {\em NAACL: Human Language Technologies}, 2018.

\bibitem{he2016deep}
Kaiming He, Xiangyu Zhang, Shaoqing Ren, and Jian Sun.
\newblock Deep residual learning for image recognition.
\newblock In {\em CVPR}, 2016.

\bibitem{hendricks2021probing}
Lisa~Anne Hendricks and Aida Nematzadeh.
\newblock Probing image-language transformers for verb understanding.
\newblock In {\em ACL-IJCNLP}, 2021.

\bibitem{hendrycks2019robustness}
Dan Hendrycks and Thomas Dietterich.
\newblock Benchmarking neural network robustness to common corruptions and
  perturbations.
\newblock In {\em ICLR}, 2019.

\bibitem{hosseinmardi2015detection}
Homa Hosseinmardi, Sabrina~Arredondo Mattson, Rahat~Ibn Rafiq, Richard Han, Qin
  Lv, and Shivakant Mishra.
\newblock Detection of cyberbullying incidents on the instagram social network.
\newblock In {\em arXiv preprint arXiv:1503.03909}, 2015.

\bibitem{hu2019evaluating}
Hexiang Hu, Ishan Misra, and Laurens van~der Maaten.
\newblock Evaluating text-to-image matching using binary image selection
  (bison).
\newblock In {\em ICCV}, 2019.

\bibitem{hu-etal-2020-systematic}
Jennifer Hu, Jon Gauthier, Peng Qian, Ethan Wilcox, and Roger Levy.
\newblock A systematic assessment of syntactic generalization in neural
  language models.
\newblock In {\em ACL}, 2020.

\bibitem{hu2021unit}
Ronghang Hu and Amanpreet Singh.
\newblock Unit: Multimodal multitask learning with a unified transformer.
\newblock In {\em arXiv preprint arXiv:2102.10772}, 2021.

\bibitem{QQPDataset}
Shankar Iyer, Nikhil Dandekar, and Kornel Csernai.
\newblock First quora dataset release: Question pairs, 2017.

\bibitem{johnson2017clevr}
Justin Johnson, Bharath Hariharan, Laurens Van Der~Maaten, Li Fei-Fei, C
  Lawrence~Zitnick, and Ross Girshick.
\newblock Clevr: A diagnostic dataset for compositional language and elementary
  visual reasoning.
\newblock In {\em CVPR}, 2017.

\bibitem{kann-etal-2019-verb}
Katharina Kann, Alex Warstadt, Adina Williams, and Samuel~R. Bowman.
\newblock Verb argument structure alternations in word and sentence embeddings.
\newblock In {\em {SC}i{L}}, 2019.

\bibitem{kiela2020hateful}
Douwe Kiela, Hamed Firooz, Aravind Mohan, Vedanuj Goswami, Amanpreet Singh,
  Pratik Ringshia, and Davide Testuggine.
\newblock The hateful memes challenge: Detecting hate speech in multimodal
  memes.
\newblock In {\em arXiv preprint arXiv:2005.04790}, 2020.

\bibitem{kim2021vilt}
Wonjae Kim, Bokyung Son, and Ildoo Kim.
\newblock Vilt: Vision-and-language transformer without convolution or region
  supervision.
\newblock In {\em ICML}, 2021.

\bibitem{kirk2021hatemoji}
Hannah~Rose Kirk, Bertram Vidgen, Paul R{\"o}ttger, Tristan Thrush, and Scott~A
  Hale.
\newblock Hatemoji: A test suite and adversarially-generated dataset for
  benchmarking and detecting emoji-based hate.
\newblock In {\em arXiv preprint arXiv:2108.05921}, 2021.

\bibitem{kocijan2020review}
Vid Kocijan, Thomas Lukasiewicz, Ernest Davis, Gary Marcus, and Leora
  Morgenstern.
\newblock A review of winograd schema challenge datasets and approaches.
\newblock In {\em arXiv preprint arXiv:2004.13831}, 2020.

\bibitem{krishna2016visual}
Ranjay Krishna, Yuke Zhu, Oliver Groth, Justin Johnson, Kenji Hata, Joshua
  Kravitz, Stephanie Chen, Yannis Kalantidis, Li-Jia Li, David~A Shamma, et~al.
\newblock Visual genome: Connecting language and vision using crowdsourced
  dense image annotations.
\newblock In {\em arXiv preprint arXiv:1602.07332}, 2016.

\bibitem{levesque2012winograd}
Hector Levesque, Ernest Davis, and Leora Morgenstern.
\newblock The winograd schema challenge.
\newblock In {\em Conference on the Principles of Knowledge Representation and
  Reasoning}, 2012.

\bibitem{li2019vsrn}
Kunpeng Li, Yulun Zhang, Kai Li, Yuanyuan Li, and Yun Fu.
\newblock Visual semantic reasoning for image-text matching.
\newblock In {\em ICCV}, 2019.

\bibitem{li2020closer}
Linjie Li, Zhe Gan, and Jingjing Liu.
\newblock A closer look at the robustness of vision-and-language pre-trained
  models.
\newblock In {\em arXiv preprint arXiv:2012.08673}, 2020.

\bibitem{li2019visualbert}
Liunian~Harold Li, Mark Yatskar, Da Yin, Cho-Jui Hsieh, and Kai-Wei Chang.
\newblock {VisualBERT: A Simple and Performant Baseline for Vision and
  Language}.
\newblock In {\em arXiv preprint arXiv:1908.03557}, 2019.

\bibitem{lin2014microsoft}
Tsung-Yi Lin, Michael Maire, Serge Belongie, James Hays, Pietro Perona, Deva
  Ramanan, Piotr Doll{\'a}r, and C~Lawrence Zitnick.
\newblock Microsoft coco: Common objects in context.
\newblock In {\em ECCV}, 2014.

\bibitem{linzen2016assessing}
Tal Linzen, Emmanuel Dupoux, and Yoav Goldberg.
\newblock Assessing the ability of lstms to learn syntax-sensitive
  dependencies.
\newblock In {\em TACL}, 2015.

\bibitem{liu-etal-2021-visually}
Fangyu Liu, Emanuele Bugliarello, Edoardo~Maria Ponti, Siva Reddy, Nigel
  Collier, and Desmond Elliott.
\newblock Visually grounded reasoning across languages and cultures.
\newblock In {\em EMNLP}, 2021.

\bibitem{lu2019vilbert}
Jiasen Lu, Dhruv Batra, Devi Parikh, and Stefan Lee.
\newblock {ViLBERT: Pretraining Task-Agnostic Visiolinguistic Representations
  for Vision-and-Language Tasks}.
\newblock In {\em NeurIPS}, 2019.

\bibitem{ordonez2011im2text}
Vicente Ordonez, Girish Kulkarni, and Tamara Berg.
\newblock Im2text: Describing images using 1 million captioned photographs.
\newblock In {\em NIPS}, 2011.

\bibitem{parcalabescu2021seeing}
Letitia Parcalabescu, Albert Gatt, Anette Frank, and Iacer Calixto.
\newblock Seeing past words: Testing the cross-modal capabilities of pretrained
  v\&l models on counting tasks.
\newblock In {\em ACL}, 2021.

\bibitem{parthasarathi-etal-2021-sometimes-want}
Prasanna Parthasarathi, Koustuv Sinha, Joelle Pineau, and Adina Williams.
\newblock Sometimes we want ungrammatical translations.
\newblock In {\em Findings of the Association for Computational Linguistics:
  EMNLP}, 2021.

\bibitem{pont-tuset2020localized-narratives}
Jordi Pont-Tuset, Jasper Uijlings, Soravit Changpinyo, Radu Soricut, and
  Vittorio Ferrari.
\newblock Connecting vision and language with localized narratives.
\newblock In {\em ECCV}, 2020.

\bibitem{radford2021clip}
Alec Radford, Jong~Wook Kim, Chris Hallacy, Aditya Ramesh, Gabriel Goh,
  Sandhini Agarwal, Girish Sastry, Amanda Askell, Pamela Mishkin, Jack Clark,
  Gretchen Krueger, and Ilya Sutskever.
\newblock Learning transferable visual models from natural language
  supervision.
\newblock In {\em ICML}, 2021.

\bibitem{rajpurkar2016squad}
Pranav Rajpurkar, Jian Zhang, Konstantin Lopyrev, and Percy Liang.
\newblock Squad: 100,000+ questions for machine comprehension of text.
\newblock In {\em arXiv preprint arXiv:1606.05250}, 2016.

\bibitem{ren2015faster}
Shaoqing Ren, Kaiming He, Ross Girshick, and Jian Sun.
\newblock Faster r-cnn: Towards real-time object detection with region proposal
  networks.
\newblock In {\em NeurIPS}, 2015.

\bibitem{rudinger2018gender}
Rachel Rudinger, Jason Naradowsky, Brian Leonard, and Benjamin Van~Durme.
\newblock Gender bias in coreference resolution.
\newblock In {\em arXiv preprint arXiv:1804.09301}, 2018.

\bibitem{sakaguchi2020winogrande}
Keisuke Sakaguchi, Ronan Le~Bras, Chandra Bhagavatula, and Yejin Choi.
\newblock Winogrande: An adversarial winograd schema challenge at scale.
\newblock In {\em AAAI}, 2020.

\bibitem{Scheuerman2019gender}
Morgan~Klaus Scheuerman, Jacob~M. Paul, and Jed~R. Brubaker.
\newblock How computers see gender: An evaluation of gender classification in
  commercial facial analysis services.
\newblock In {\em ACM: Human Computer Interaction}, 2019.

\bibitem{sennrich2015neural}
Rico Sennrich, Barry Haddow, and Alexandra Birch.
\newblock Neural machine translation of rare words with subword units.
\newblock In {\em arXiv preprint arXiv:1508.07909}, 2015.

\bibitem{sharma2018conceptual}
Piyush Sharma, Nan Ding, Sebastian Goodman, and Radu Soricut.
\newblock Conceptual captions: A cleaned, hypernymed, image alt-text dataset
  for automatic image captioning.
\newblock In {\em ACL}, 2018.

\bibitem{shekhar2017foil}
Ravi Shekhar, Sandro Pezzelle, Yauhen Klimovich, Aurelie Herbelot, Moin Nabi,
  Enver Sangineto, and Raffaella Bernardi.
\newblock "foil it! find one mismatch between image and language caption".
\newblock In {\em ACL}, 2017.

\bibitem{sidorov2020textcaps}
Oleksii Sidorov, Ronghang Hu, Marcus Rohrbach, and Amanpreet Singh.
\newblock Textcaps: a dataset for image captioning with reading comprehension.
\newblock In {\em ECCV}, 2020.

\bibitem{simonyan2015very}
Karen Simonyan and Andrew Zisserman.
\newblock Very deep convolutional networks for largescale image recognition.
\newblock In {\em CVPR}, 2015.

\bibitem{singh2020we}
Amanpreet Singh, Vedanuj Goswami, and Devi Parikh.
\newblock Are we pretraining it right? digging deeper into visio-linguistic
  pretraining.
\newblock In {\em arXiv preprint arXiv:2004.08744}, 2020.

\bibitem{singh2022flava}
Amanpreet Singh, Ronghang Hu, Vedanuj Goswami, Guillaume Couairon, Wojciech
  Galuba, Marcus Rohrbach, and Douwe Kiela.
\newblock Flava: A foundational language and vision alignment model.
\newblock In {\em CVPR}, 2022.

\bibitem{singh2019towards}
Amanpreet Singh, Vivek Natarajan, Meet Shah, Yu Jiang, Xinlei Chen, Dhruv
  Batra, Devi Parikh, and Marcus Rohrbach.
\newblock Towards vqa models that can read.
\newblock In {\em CVPR}, 2019.

\bibitem{sinha2021matterslittle}
Koustuv Sinha, Robin Jia, Dieuwke Hupkes, Joelle Pineau, Adina Williams, and
  Douwe Kiela.
\newblock Masked language modeling and the distributional hypothesis: Order
  word matters pre-training for little.
\newblock In {\em EMNLP}, 2021.

\bibitem{sinha-etal-2021-unnatural}
Koustuv Sinha, Prasanna Parthasarathi, Joelle Pineau, and Adina Williams.
\newblock {UnNatural} {L}anguage {I}nference.
\newblock In {\em ACL-IJCNLP}, 2021.

\bibitem{Socher2013RecursiveDM}
Richard Socher, Alex Perelygin, Jean Wu, Jason Chuang, Christopher~D. Manning,
  A. Ng, and Christopher Potts.
\newblock Recursive deep models for semantic compositionality over a sentiment
  treebank.
\newblock In {\em EMNLP}, 2013.

\bibitem{srinivasan2021wit}
Krishna Srinivasan, Karthik Raman, Jiecao Chen, Michael Bendersky, and Marc
  Najork.
\newblock Wit: Wikipedia-based image text dataset for multimodal multilingual
  machine learning.
\newblock In {\em arXiv preprint arXiv:2103.01913}, 2021.

\bibitem{suhr2017corpus}
Alane Suhr, Mike Lewis, James Yeh, and Yoav Artzi.
\newblock A corpus of natural language for visual reasoning.
\newblock In {\em ACL}, 2017.

\bibitem{suryawanshi2021trollmeme}
Shardul Suryawanshi and Bharathi~Raja Chakravarthi.
\newblock Findings of the shared task on troll meme classification in {T}amil.
\newblock In {\em Proceedings of the First Workshop on Speech and Language
  Technologies for Dravidian Languages}, 2021.

\bibitem{tan2020lxmert}
Hao Tan and Mohit Bansal.
\newblock Lxmert: Learning cross-modality encoder representations from
  transformers.
\newblock In {\em EMNLP-IJCNLP}, 2020.

\bibitem{thomee2016yfcc100m}
Bart Thomee, David~A Shamma, Gerald Friedland, Benjamin Elizalde, Karl Ni,
  Douglas Poland, Damian Borth, and Li-Jia Li.
\newblock Yfcc100m: The new data in multimedia research.
\newblock In {\em Communications of the ACM}, 2016.

\bibitem{thrush2020investigating}
Tristan Thrush, Ethan Wilcox, and Roger Levy.
\newblock Investigating novel verb learning in {BERT}: Selectional preference
  classes and alternation-based syntactic generalization.
\newblock In {\em Proceedings of the Third BlackboxNLP Workshop on Analyzing
  and Interpreting Neural Networks for NLP}, 2020.

\bibitem{vaswani2017attention}
Ashish Vaswani, Noam Shazeer, Niki Parmar, Jakob Uszkoreit, Llion Jones,
  Aidan~N Gomez, {\L}ukasz Kaiser, and Illia Polosukhin.
\newblock Attention is all you need.
\newblock In {\em NeurIPS}, 2017.

\bibitem{vedantam2021curi}
Ramakrishna Vedantam, Arthur Szlam, Maximillian Nickel, Ari Morcos, and
  Brenden~M Lake.
\newblock Curi: A benchmark for productive concept learning under uncertainty.
\newblock In {\em ICML}, 2021.

\bibitem{warstadt-etal-2019-investigating}
Alex Warstadt, Yu Cao, Ioana Grosu, Wei Peng, Hagen Blix, Yining Nie, Anna
  Alsop, Shikha Bordia, Haokun Liu, Alicia Parrish, Sheng-Fu Wang, Jason Phang,
  Anhad Mohananey, Phu~Mon Htut, Paloma Jeretic, and Samuel~R. Bowman.
\newblock Investigating {BERT}{'}s knowledge of language: Five analysis methods
  with {NPI}s.
\newblock In {\em EMNLP-IJCNLP}, 2019.

\bibitem{warstadt-etal-2020-blimp-benchmark}
Alex Warstadt, Alicia Parrish, Haokun Liu, Anhad Mohananey, Wei Peng, Sheng-Fu
  Wang, and Samuel~R. Bowman.
\newblock {BL}i{MP}: The benchmark of linguistic minimal pairs for {E}nglish.
\newblock In {\em TACL}, 2020.

\bibitem{williams-etal-2018-latent}
Adina Williams, Andrew Drozdov, and Samuel~R. Bowman.
\newblock Do latent tree learning models identify meaningful structure in
  sentences?
\newblock In {\em TACL}, 2018.

\bibitem{williams2017broad}
Adina Williams, Nikita Nangia, and Samuel~R Bowman.
\newblock A broad-coverage challenge corpus for sentence understanding through
  inference.
\newblock In {\em arXiv preprint arXiv:1704.05426}, 2017.

\bibitem{winograd1972understanding}
Terry Winograd.
\newblock Understanding natural language.
\newblock In {\em Cognitive psychology}, 1972.

\bibitem{wolf-etal-2020-transformers}
Thomas Wolf, Lysandre Debut, Victor Sanh, Julien Chaumond, Clement Delangue,
  Anthony Moi, Pierric Cistac, Tim Rault, Rémi Louf, Morgan Funtowicz, Joe
  Davison, Sam Shleifer, Patrick von Platen, Clara Ma, Yacine Jernite, Julien
  Plu, Canwen Xu, Teven~Le Scao, Sylvain Gugger, Mariama Drame, Quentin Lhoest,
  and Alexander~M. Rush.
\newblock Transformers: State-of-the-art natural language processing.
\newblock In {\em EMNLP: System Demonstrations}, 2020.

\bibitem{xie2018visual}
Ning Xie, Farley Lai, Derek Doran, and Asim Kadav.
\newblock Visual entailment task for visually-grounded language learning.
\newblock In {\em arXiv preprint arXiv:1811.10582}, 2018.

\bibitem{young2014image}
Peter Young, Alice Lai, Micah Hodosh, and Julia Hockenmaier.
\newblock From image descriptions to visual denotations: New similarity metrics
  for semantic inference over event descriptions.
\newblock In {\em TACL}, 2014.

\bibitem{zellers2019recognition}
Rowan Zellers, Yonatan Bisk, Ali Farhadi, and Yejin Choi.
\newblock From recognition to cognition: Visual commonsense reasoning.
\newblock In {\em CVPR}, 2019.

\bibitem{zhang2021vinvl}
Pengchuan Zhang, Xiujun Li, Xiaowei Hu, Jianwei Yang, Lei Zhang, Lijuan Wang,
  Yejin Choi, and Jianfeng Gao.
\newblock Vinvl: Revisiting visual representations in vision-language models.
\newblock In {\em CVPR}, 2021.

\bibitem{zhao2018gender}
Jieyu Zhao, Tianlu Wang, Mark Yatskar, Vicente Ordonez, and Kai-Wei Chang.
\newblock Gender bias in coreference resolution: Evaluation and debiasing
  methods.
\newblock In {\em arXiv preprint arXiv:1804.06876}, 2018.

\bibitem{zhong2016content}
Haoti Zhong, Hao Li, Anna~Cinzia Squicciarini, Sarah~Michele Rajtmajer,
  Christopher Griffin, David~J Miller, and Cornelia Caragea.
\newblock Content-driven detection of cyberbullying on the instagram social
  network.
\newblock In {\em IJCAI}, 2016.

\bibitem{zhu2016visual7w}
Yuke Zhu, Oliver Groth, Michael Bernstein, and Li Fei-Fei.
\newblock Visual7w: Grounded question answering in images.
\newblock In {\em CVPR}, 2016.

\end{thebibliography}
}
\end{document}




\appendix

\section{Confidence Intervals}

We provide confidence intervals for the overall model results on Winoground. We divided the dataset into 4 groups of equal size to get 4 scores for each model and score-type, and used Student's t-distribution to compute the confidence intervals.

\begin{table}[H]
\centering
\begin{tabular}{l|rr|rr|rr}
\toprule
 \textit{Model}      & \textit{Text} &          & \textit{Image} &         & \textit{Group} &         \\
\midrule
 MTurk Human                  & \textbf{89.50} & [\textbf{80.83},\textbf{98.17}] & \textbf{88.50} & [\textbf{79.00},\textbf{98.00}] & \textbf{85.50} & [\textbf{73.84},\textbf{97.16}] \\\midrule
 VinVL                        & \textbf{37.75} & [\textbf{28.71},\textbf{46.79}] & 17.75 & [11.21,24.29] & 14.50 & [6.65,\textbf{22.35}]  \\
 UNITER$_{large}$             & \textbf{38.00} & [\textbf{33.32},\textbf{42.68}] & 14.00 & [6.77,21.23]  & 10.50 & [8.45,12.55]  \\
 UNITER$_{base}$              & \textbf{32.25} & [\textbf{25.84},\textbf{38.66}] & 13.25 & [7.68,18.82]  & 10.00 & [7.75,12.25]  \\
 ViLLA$_{large}$              & \textbf{37.00} & [\textbf{31.05},\textbf{42.95}] & 13.25 & [7.83,18.67]  & 11.00 & [7.10,14.90]  \\
 ViLLA$_{base}$               & \textbf{30.00} & [\textbf{25.32},\textbf{34.68}] & 12.00 & [8.33,15.67]  & 8.00 & [5.75,10.25]   \\
 VisualBERT$_{base}$          & 15.50 & [9.34,21.66]  & 2.50 & [0.00,6.29]    & 1.50 & [0.00,3.55]    \\
 ViLT (ViT-B/32)              & \textbf{34.75} & [\textbf{29.03},\textbf{40.47}] & 14.00 & [8.49,19.51]  & 9.25 & [6.53,11.97]   \\
 LXMERT                       & 19.25 & [16.53,21.97] & 7.00 & [3.10,10.90]   & 4.00 & [2.70,5.30]    \\
 ViLBERT$_{base}$             & 23.75 & [18.03,\textbf{29.47}] & 7.25 & [3.97,10.53]   & 4.75 & [1.47,8.03]    \\
 UniT$_{ITM Finetuned}$       & 19.50 & [14.73,24.27] & 6.25 & [0.53,11.97]   & 4.00 & [2.70,5.30]    \\
 FLAVA$_{ITM}$                & \textbf{32.25} & [20.04,\textbf{44.46}] & 20.50 & [14.34,\textbf{26.66}] & 14.25 & [8.53,\textbf{19.97}]  \\
 FLAVA$_{Contrastive}$        & \textbf{25.25} & [19.99,\textbf{30.51}] & 13.50 & [8.55,18.45]  & 9.00 & [5.10,12.90]   \\
 CLIP (ViT-B/32)              & \textbf{30.75} & [\textbf{25.03},\textbf{36.47}] & 10.50 & [6.29,14.71]  & 8.00 & [4.56,11.44]   \\
 VSE++$_{COCO}$ (ResNet)      & 22.75 & [19.22,\textbf{26.28}] & 8.00 & [6.70,9.30]    & 4.00 & [1.40,6.60]    \\
 VSE++$_{COCO}$ (VGG)         & 18.75 & [17.23,20.27] & 5.50 & [3.45,7.55]    & 3.50 & [2.58,4.42]    \\
 VSE++$_{Flickr30k}$ (ResNet) & 20.00 & [12.77,\textbf{27.23}] & 5.00 & [0.89,9.11]    & 2.75 & [0.75,4.75]    \\
 VSE++$_{Flickr30k}$ (VGG)    & 19.75 & [14.49,\textbf{25.01}] & 6.25 & [2.27,10.23]   & 4.50 & [2.91,6.09]    \\
 VSRN$_{COCO}$                & 17.50 & [9.54,\textbf{25.46}]  & 7.00 & [1.19,12.81]   & 3.75 & [0.00,8.50]    \\
 VSRN$_{Flickr30k}$           & 20.00 & [13.25,\textbf{26.75}] & 5.00 & [2.09,7.91]    & 3.50 & [2.58,4.42]    \\
\bottomrule
\end{tabular}
\caption{95\% confidence intervals for the aggregate results on Winoground. Results above chance are shown in \textbf{bold}.}
\label{tab:results-aggr}
\end{table}

\newpage

\section{Impact of Pretraining Data Size and Model Type on Model Performance}
Correlations between pretraining data size and model performance are highly significant in every case and the numbers are shown in the main paper. We show plots in the figures below. Most of the single-stream models perform slightly above chance on the text score. CLIP and FLAVA are the only dual-stream models which perform above chance, and they have drastically more training data than all other models.

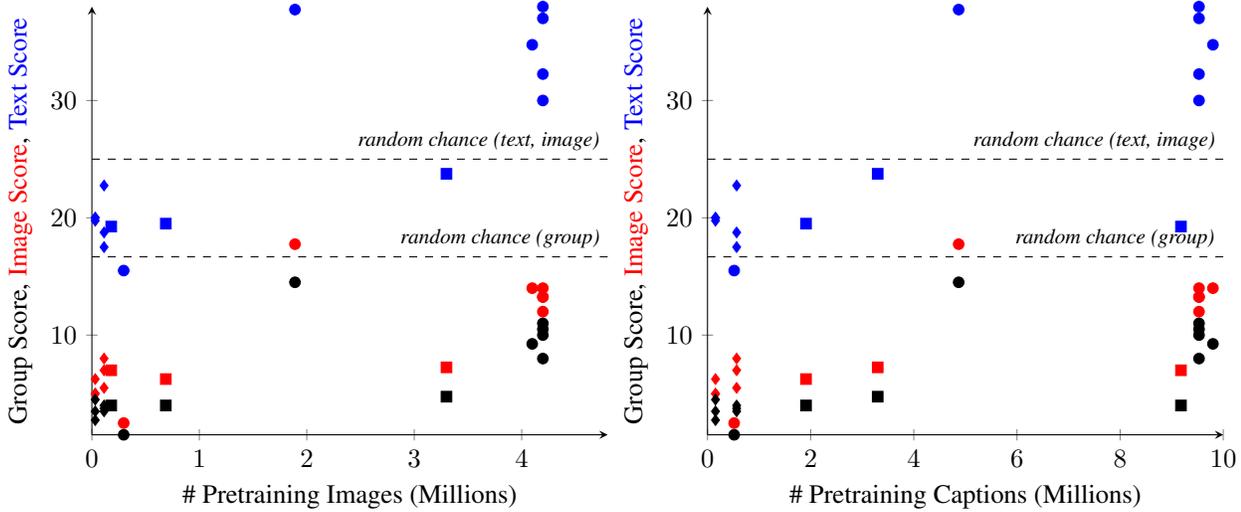
\begin{figure}[H]%
    \centering
    \hspace*{-1em}\begin{tikzpicture}[scale=1]
    \begin{axis}[%
        scatter/classes={%
            textrnn={blue, mark=diamond*}, textdual={blue, mark=square*}, textsingle={blue}, imgrnn={red, mark=diamond*}, imgdual={red, mark=square*}, imgsingle={red}, grprnn={mark=diamond*}, grpdual={mark=square*}, grpsingle={}, lbl={opacity=0}},
            axis x line=bottom,
            axis y line=left,
            xlabel = \# Pretraining Images (Millions),
            ylabel = {Group Score, {\color{red} Image Score}, {\color{blue} Text Score}}
            ],
        \addplot[
            scatter,
            only marks,%
            scatter src=explicit symbolic,
        ]
            table[meta=class] {
            x y class label
            4.197 32.25 textsingle     UNITER$_{base}$
            4.197 30.0 textsingle      VILLA$_{base}$
            4.197 38 textsingle     UNITER$_{large}$
            4.197 37 textsingle      VILLA$_{large}$
            1.89 37.75 textsingle       VinVL$_{base}$
            0.296 15.5 textsingle     VisualBERT
            4.098 34.75 textsingle      ViLT
            0.18 19.25 textdual       LXMERT
            3.3 23.75 textdual        ViLBERT
            .688 19.5 textdual      UniT
            .113 22.75 textrnn VSE++$_{COCO}$ (ResNet)
            .113 18.75 textrnn VSE++$_{COCO}$ (VGG)
            .031 20.00 textrnn VSE++$_{F30k}$ (ResNet)
            .031 19.75 textrnn VSE++$_{F30k}$ (VGG)
            .113 17.50 textrnn VRSN$_{COCO}$
            .031 20 textrnn VRSN$_{F30k}$
        
            4.197 13.25 imgsingle     UNITER$_{base}$
            4.197 12 imgsingle      VILLA$_{base}$
            4.197 14 imgsingle      UNITER$_{large}$
            4.197 13.25 imgsingle     VILLA$_{large}$
            1.89 17.75 imgsingle       VinVL$_{base}$
            .296 2.5 imgsingle       VisualBERT
            4.098 14 imgsingle    ViLT
            0.18 7 imgdual       LXMERT
            3.3 7.25 imgdual         ViLBERT
            .688 6.25 imgdual          UniT
            .113 8.0 imgrnn VSE++$_{COCO}$ (ResNet)
            .113 5.50 imgrnn VSE++$_{COCO}$ (VGG)
            .031 5.00 imgrnn VSE++$_{F30k}$ (ResNet)
            .031 6.25 imgrnn VSE++$_{F30k}$ (VGG)
            .113 7.00 imgrnn VRSN$_{COCO}$
            .031 5.00 imgrnn VRSN$_{F30k}$
            
             1.89 14.50 grpsingle VinVL$_{base}$
            4.197 10.50 grpsingle UNITER$_{large}$
            4.197  10.00 grpsingle UNITER$_{base}$
            4.197  11.00 grpsingle VILLA$_{large}$
            4.197  8.00 grpsingle VILLA$_{base}$
            .296 1.50 grpsingle VisualBERT
            4.098 9.25 grpsingle ViLT
            0.18 4.00 grpdual LXMERT
            3.3 4.75 grpdual  ViLBERT
            .688 4.00 grpdual  UniT
            .113 4.00 grprnn VSE++$_{COCO}$ (ResNet)
            .113 3.50 grprnn VSE++$_{COCO}$ (VGG)
            .031 2.75 grprnn VSE++$_{F30k}$ (ResNet)
            .031 4.50 grprnn VSE++$_{F30k}$ (VGG)
            .113 3.75 grprnn VRSN$_{COCO}$
            .031 3.50 grprnn VRSN$_{F30k}$
            };
        
               \addplot[
            scatter,
            only marks,%
            scatter src=explicit symbolic,
            nodes near coords*={\Label},
            every node near coord/.append style={font=\footnotesize},
            visualization depends on={value \thisrow{label} \as \Label},
        ]
            table[meta=class] {
                x y class label
                3.6 25 lbl \textit{random chance (text, image)}
                3.8 16.67 lbl \textit{random chance (group)}
            };

        
       \addplot[mark=none, black, dashed, label={random chance}] coordinates {(0,25) (4.8,25)};
       \addplot[mark=none, black, dashed, label={random chance}] coordinates {(0,16.67) (4.8,16.67)};
           
    \end{axis}
\end{tikzpicture}
\begin{tikzpicture}[scale=1]
    \begin{axis}[%
        scatter/classes={%
            textrnn={blue, mark=diamond*}, textdual={blue, mark=square*}, textsingle={blue}, imgrnn={red, mark=diamond*}, imgdual={red, mark=square*}, imgsingle={red}, grprnn={mark=diamond*}, grpdual={mark=square*}, grpsingle={}, lbl={opacity=0}},
            axis x line=bottom,
            axis y line=left,
            xlabel = \# Pretraining Captions (Millions),
            ylabel = {Group Score, {\color{red} Image Score}, {\color{blue} Text Score}}
            ],
        \addplot[
            scatter,
            only marks,%
            scatter src=explicit symbolic,
        ]
            table[meta=class] {
                x y class label
                9.53 32.25 textsingle UNITER$_{base}$
                9.53 30 textsingle VILLA$_{base}$
                9.53 38 textsingle UNITER$_{large}$
                9.53 37 textsingle VILLA$_{large}$
                4.87 37.75 textsingle VinVL$_{base}$
                0.517 15.5 textsingle VisualBERT
                9.8 34.75 textsingle ViLT
                9.18 19.25 textdual LXMERT
                3.3 23.75 textdual ViLBERT
                1.91 19.5 textdual UniT
                .565 22.75 textrnn VSE++$_{COCO}$ (ResNet)
                .565 18.75 textrnn VSE++$_{COCO}$ (VGG)
                .155 20.00 textrnn VSE++$_{F30k}$ (ResNet)
                .155 19.75 textrnn VSE++$_{F30k}$ (VGG)
                .565 17.50 textrnn VRSN$_{COCO}$
                .155 20 textrnn VRSN$_{F30k}$
                
                9.53 13.25 imgsingle UNITER$_{base}$
                9.53 12 imgsingle VILLA$_{base}$
                9.53 14 imgsingle UNITER$_{large}$
                9.53 13.25 imgsingle VILLA$_{large}$
                4.87 17.75 imgsingle VinVL$_{base}$
                .517 2.5 imgsingle VisualBERT
                9.8 14 imgsingle ViLT
                9.18 7 imgdual LXMERT
                3.3 7.25 imgdual ViLBERT
                1.91 6.25 imgdual UniT
                .565 8.0 imgrnn VSE++$_{COCO}$ (ResNet)
                .565 5.50 imgrnn VSE++$_{COCO}$ (VGG)
                .155 5.00 imgrnn VSE++$_{F30k}$ (ResNet)
                .155 6.25 imgrnn VSE++$_{F30k}$ (VGG)
                .565 7.00 imgrnn VRSN$_{COCO}$
                .155 5.00 imgrnn VRSN$_{F30k}$
                
                4.87 14.50 grpsingle VinVL$_{base}$
                9.53 10.50 grpsingle UNITER$_{large}$
                9.53 10.00 grpsingle UNITER$_{base}$
                9.53 11.00 grpsingle VILLA$_{large}$
                9.53 8.00 grpsingle VILLA$_{base}$
                .517 1.50 grpsingle VisualBERT
                9.8 9.25 grpsingle ViLT
                9.18 4.00 grpdual LXMERT
                3.3 4.75 grpdual  ViLBERT
                1.91 4.00 grpdual  UniT
                .565 4.00 grprnn VSE++$_{COCO}$ (ResNet)
                .565 3.50 grprnn VSE++$_{COCO}$ (VGG)
                .155 2.75 grprnn VSE++$_{F30k}$ (ResNet)
                .155 4.50 grprnn VSE++$_{F30k}$ (VGG)
                .565 3.75 grprnn VRSN$_{COCO}$
                .155 3.50 grprnn VRSN$_{F30k}$

            };
            
             \addplot[
            scatter,
            only marks,%
            scatter src=explicit symbolic,
            nodes near coords*={\Label},
            every node near coord/.append style={font=\footnotesize},
            visualization depends on={value \thisrow{label} \as \Label},
        ]
            table[meta=class] {
                x y class label
                7.5 25 lbl \textit{random chance (text, image)}
                7.9 16.67 lbl \textit{random chance (group)}
            };

        
       \addplot[mark=none, black, dashed, label={random chance}] coordinates {(0,25) (10,25)};
       \addplot[mark=none, black, dashed, label={random chance}] coordinates {(0,16.67) (10,16.67)};
    \end{axis}
\end{tikzpicture}
    \caption{Graphs of the model performance on Winoground for each model by the number of pretraining images (left) and pretraining captions (right). $\Diamond=$ dual-stream RNNs, $\square=$ dual-stream transformers, $\bigcircle=$ single-stream transformers. CLIP and FLAVA are removed as outliers. Backbone pretraining data is not included.}%
    \label{fig:teaser-ex}%
\end{figure}

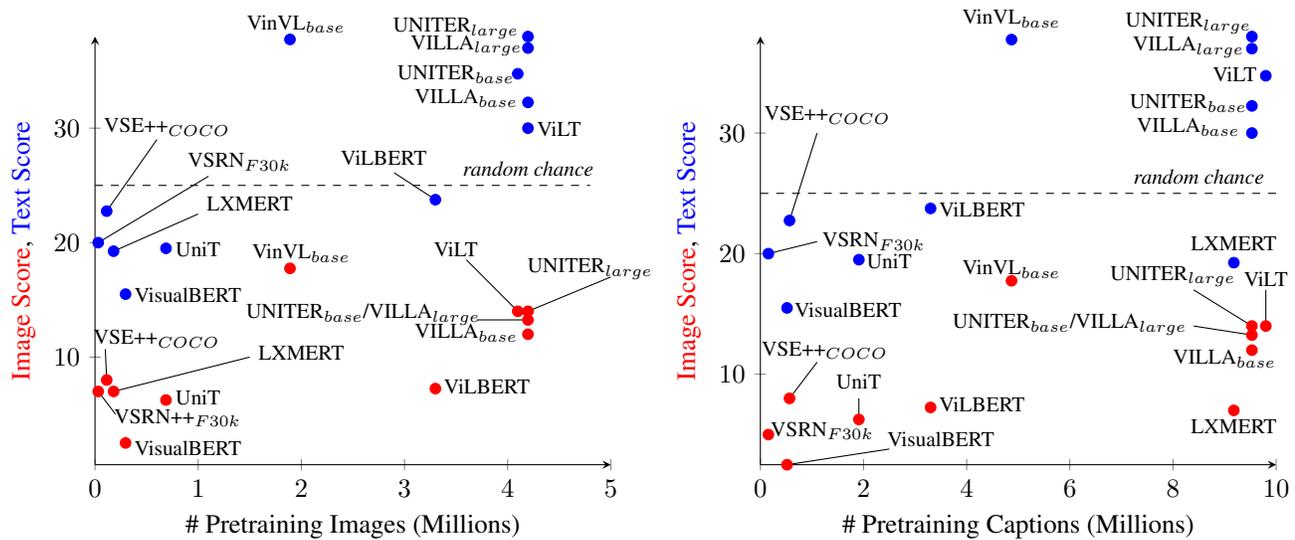
\begin{figure}[H]
    \centering
    \hspace*{-1em}\begin{tikzpicture}[scale=1]
    \begin{axis}[%
        scatter/classes={%
            text={blue}, img={red}, lbl={opacity=0}},
            axis x line=bottom,
            axis y line=left,
            xlabel = \# Pretraining Images (Millions),
            ylabel = {{\color{red} Image Score}, {\color{blue} Text Score}}
            ],
        \addplot[
            scatter,
            only marks,%
            scatter src=explicit symbolic,
        ]
            table[meta=class] {
            x y class label
            4.197 32.25 text     UNITER$_{base}$
            4.197 30.0 text      VILLA$_{base}$
            4.197 38 text     UNITER$_{large}$
            4.197 37 text      VILLA$_{large}$
            1.89 37.75 text       VinVL$_{base}$
            0.296 15.5 text     VisualBERT
            4.098 34.75 text      ViLT
            0.18 19.25 text       LXMERT
            3.3 23.75 text        ViLBERT
            .688 19.5 text      UniT
            .113 22.75 text       VSE++$_{COCO}$
            .031 20 text         VSRN$_{Flickr30k}$
            
            4.197 13.25 img     UNITER$_{base}$
            4.197 12 img     VILLA$_{base}$
            4.197 14 img     UNITER$_{large}$
            1.89 17.75 img       VinVL$_{base}$
            .296 2.5 img       VisualBERT
            4.098 14 img    ViLT
            0.18 7 img       LXMERT
            3.3 7.25 img         ViLBERT
            .688 6.25 img          UniT
            .113 8.0 img        VSE++$_{COCO}$
            .031 7 img       VRSN$_{Flickr30k}$
            };
        
        \addplot[
            scatter,
            only marks,%
            scatter src=explicit symbolic,
            nodes near coords*={\Label},
            every node near coord/.append style={font=\footnotesize},
            visualization depends on={value \thisrow{label} \as \Label},
        ]
            table[meta=class] {
                x y class label
                3.5 33.1 lbl     UNITER$_{base}$
                3.6 31 lbl      VILLA$_{base}$
                3.5 37 lbl     UNITER$_{large}$
                3.6 35.5 lbl      VILLA$_{large}$
                1.95 37.5 lbl       VinVL$_{base}$
                0.9 14 lbl     VisualBERT
                4.5 28.5 lbl      ViLT
                1.5 22 lbl       LXMERT
                2.8 26 lbl        ViLBERT
                1 18 lbl      UniT
                .7 28.5 lbl       VSE++$_{COCO}$
                1.4 25.5 lbl         VSRN$_{F30k}$
                
                2.6 12 lbl     UNITER$_{base}$/VILLA$_{large}$
                3.6 10.5 lbl     VILLA$_{base}$
                4.8 16 lbl     UNITER$_{large}$
                2 17.5 lbl       VinVL$_{base}$
                .9 0.6 lbl       VisualBERT
                3.5 18 lbl    ViLT
                2 9 lbl       LXMERT
                3.8 6 lbl         ViLBERT
                1 5 lbl          UniT
                .6 10 lbl        VSE++$_{COCO}$
                .8 3.1 lbl       VSRN++$_{F30k}$
                
                4.2 25 lbl \textit{random chance}
            };
            \addplot[] coordinates{
                (1, 22.5)
                (0.18, 19.25)
            };
            \addplot[] coordinates {
            	(1.5, 10)
            	(0.18, 7)
            };
            \addplot[] coordinates {
            	(.1, 10.7)
            	(.113, 8.0)
            };
            \addplot[] coordinates{
                (.6, 29)
                (.113, 22.75)
            };
            \addplot[] coordinates{
                (1.1, 25.8)
                (.031, 20)
            };
            \addplot[] coordinates{
                (0.2, 5)
                (.031, 7)
            };
            \addplot[] coordinates{
                (3.6, 18.5)
                (4.098, 14)
            };
            \addplot[] coordinates{
                (5, 17)
                (4.197, 14)
            };
            \addplot[] coordinates{
                (3.7, 13.25)
                (4.197, 13.25)
            };
            \addplot[] coordinates{
                (2.9, 26.1)
                (3.3, 23.75)
            };
           \addplot[mark=none, black, dashed, label={random chance}] coordinates {(0,25) (4.8,25)};
           
    \end{axis}
\end{tikzpicture}
\begin{tikzpicture}[scale=1]
    \begin{axis}[%
        scatter/classes={%
            text={blue}, img={red}, lbl={opacity=0}},
            axis x line=bottom,
            axis y line=left,
            xlabel = \# Pretraining Captions (Millions),
            ylabel = {{\color{red} Image Score}, {\color{blue} Text Score}}
            ],
        \addplot[
            scatter,
            only marks,%
            scatter src=explicit symbolic,
        ]
            table[meta=class] {
                x y class label
                9.53 32.25 text UNITER$_{base}$
                9.53 30 text VILLA$_{base}$
                9.53 38 text UNITER$_{large}$
                9.53 37 text VILLA$_{large}$
                4.87 37.75 text VinVL$_{base}$
                0.517 15.5 text VisualBERT
                9.8 34.75 text ViLT
                9.18 19.25 text LXMERT
                3.3 23.75 text ViLBERT
                1.91 19.5 text UniT
                .565 22.75 text VSE++$_{COCO}$
                .155 20 text VRSN$_{F30k}$
                
                9.53 13.25 img UNITER$_{base}$
                9.53 12 img VILLA$_{base}$
                9.53 14 img UNITER$_{large}$
                4.87 17.75 img VinVL$_{base}$
                .517 2.5 img VisualBERT
                9.8 14 img ViLT
                9.18 7 img LXMERT
                3.3 7.25 img ViLBERT
                1.91 6.25 img UniT
                .565 8.0 img VSE++$_{COCO}$
                .155 5 img VSRN$_{F30k}$
                
            };
            
        \addplot[
            scatter,
            only marks,%
            scatter src=explicit symbolic,
            nodes near coords*={\Label},
            every node near coord/.append style={font=\footnotesize},
            visualization depends on={value \thisrow{label} \as \Label},
        ]
            table[meta=class] {
                x y class label
                8.3 31 lbl UNITER$_{base}$
                8.3 29 lbl VILLA$_{base}$
                8.3 37.5 lbl UNITER$_{large}$
                8.3 35.5 lbl VILLA$_{large}$
                4.87 38.0 lbl VinVL$_{base}$
                1.7 14 lbl VisualBERT
                9.2 33.5 lbl ViLT
                9.18 19.5 lbl LXMERT
                4.3 22.3 lbl ViLBERT
                2.5 18 lbl UniT
                1.3 30 lbl VSE++$_{COCO}$
                2.2 19.5 lbl VSRN$_{F30k}$
                
                6 12.5 lbl UNITER$_{base}$/VILLA$_{large}$
                9 9.8 lbl VILLA$_{base}$
                8 16.5 lbl UNITER$_{large}$
                4.87 17.2 lbl VinVL$_{base}$
                3.5 3.25 lbl VisualBERT
                9.8 16.5 lbl ViLT
                9.18 4.3 lbl LXMERT
                4.3 6.1 lbl ViLBERT
                1.91 8 lbl UniT
                1.25 10.5 lbl VSE++$_{COCO}$
                1.2 3.8 lbl VSRN$_{F30k}$
                
                8.5 25 lbl \textit{random chance}
            };
            
            \addplot[] coordinates{
                (.565, 22.75)
                (1.1, 30)
            };
            \addplot[] coordinates{
                (1.1, 21)
                (.155, 20)
            };
            \addplot[] coordinates{
                (.517, 2.5)
                (2.5, 3.8)
            };
            \addplot[] coordinates{
                (.565, 8.0)
                (1.1, 11)
            };
            \addplot[] coordinates{
                (1.91, 6.25)
                (1.91, 8)
            };
            \addplot[] coordinates{
                (8.4, 14)
                (9.53, 13.25)
            };
            \addplot[] coordinates{
                (8.4, 17)
                (9.53, 14)
            };
            \addplot[] coordinates{
                (9.8, 17)
                (9.8, 14)
            };
       \addplot[mark=none, black, dashed] coordinates {(0,25) (10,25)};
    \end{axis}
\end{tikzpicture}
    \caption{Graphs of the model performance on Winoground for each model by the number of pretraining images (left) and pretraining captions (right). This is a finer-grained version of \cref{tab:results-aggr}, with model names instead of grouping by architecture; we again exclude CLIP and FLAVA as their pretraining dataset sizes are outliers. We only show the best VSE++ and VSRN configurations and do not show group scores due to clutter issues.}
    \label{fig:my_label}
\end{figure}

\newpage

\section{Linguistic Tag Breakdown}

This section reports every different swap-dependent linguistic tag that our annotators gave examples. Many of these fine-grained linguistic tags are used for multiple examples, although some tags are only used once in the dataset.

\begin{table}[H]
\centering
\resizebox{\textwidth}{!}{%
\begin{tabular}{lll}
\toprule
 Tag                                & Fine-Grained Tag                                       & Example                                                                                             \\
 \rowcolor{LightBlue} \midrule     & Noun Phrase, Determiner-Numeral                               & [a person] carrying [more than one flotation device]                                                                          \\
 \rowcolor{LighterBlue}           & Noun Phrase                                                   & [a person] holding up [books]                                                                                                 \\
 \rowcolor{LightBlue}             & Determiner-Numeral, Noun Phrase                               & [a lightbulb] surrounding [some plants]                                                                                       \\
 \rowcolor{LighterBlue} Object    & Noun Phrase, Determiner-Possessive                            & [a deer's nose] is resting on [a child's hand]                                                                                \\
 \rowcolor{LightBlue}             & Noun Phrase, Adjective-Color                                  & aerial view of a green tree in [the brown freshly turned soil] next to [a green field]                                        \\
 \rowcolor{LighterBlue}           & Pronoun, Noun Phrase                                          & [the person] wears a hat but [it] doesn't                                                                                     \\
 \rowcolor{LightBlue}             & Determiner-Numeral Phrase                                     & [one] is in a boat and [almost everyone] is swimming                                                                          \\
 \rowcolor{LighterBlue}           & Pronoun, Verb-Intransitive                                    & [it] ran away while [they] pursued                                                                                            \\
 \rowcolor{LightBlue}             & Noun                                                          & more [bicycles] than [cars]                                                                                                   \\
 \rowcolor{LightGreen} \midrule     & Adjective-Age                                                 & [an older] person blocking [a younger] person                                                                                 \\
 \rowcolor{LighterGreen}          & Scope, Preposition                                            & racing [over] it []                                                                                                           \\
 \rowcolor{LightGreen}            & Verb-Intransitive, Verb-Transitive Phrase                     & a kid [threw a basketball] then [jumped]                                                                                      \\
 \rowcolor{LighterGreen}          & Verb-Intransitive, Adjective-Manner                           & the younger person is [making noise] while the other is [silent]                                                              \\
 \rowcolor{LightGreen}            & Negation, Noun Phrase, Preposition Phrase                     & a person [with long braids] is exercising in front of a person [without braids]                                               \\
 \rowcolor{LighterGreen}          & Scope, Preposition, Verb-Intransitive                         & [out]1[swam]2 the person in the red swimcap []2[]1                                                                            \\
 \rowcolor{LightGreen}            & Noun Phrase, Adjective-Animate                                & the one on the left is [sad] and the other is [happy]                                                                         \\
 \rowcolor{LighterGreen}          & Adjective-Size                                                & the [taller] person hugs the [shorter] person                                                                                 \\
 \rowcolor{LightGreen}            & Determiner-Possessive                                         & the [person's] leg is on the [dog's] torso                                                                                    \\
 \rowcolor{LighterGreen}          & Adjective-Texture                                             & [smooth] shoes are on a [soft] floor                                                                                          \\
 \rowcolor{LightGreen}            & Adjective-Color                                               & painting the [white] wall [red]                                                                                               \\
 \rowcolor{LighterGreen}          & Scope                                                         & [getting] a horse [] wet                                                                                                      \\
 \rowcolor{LightGreen}            & Preposition Phrase                                            & flat [at the bottom] and pointy [on top]                                                                                      \\
 \rowcolor{LighterGreen}          & Relative Clause, Scope                                        & the person [who is wearing a crown] is kissing a frog []                                                                      \\
 \rowcolor{LightGreen}            & Adjective-Height                                              & a [taller] person wearing blue standing next to a [shorter] person                                                            \\
 \rowcolor{LighterGreen}          & Verb-Intransitive Phrase, Preposition                         & the gesture of the person [sitting down] is supporting the understanding of the person [standing up]                          \\
 \rowcolor{LightGreen}            & Verb-Intransitive, Determiner-Numeral                         & some people are [standing] but more are [sitting]                                                                             \\
 \rowcolor{LighterGreen}          & Determiner-Numeral                                            & [one]1 person[]2 wearing [two]1 scarf[s]2                                                                                     \\
 \rowcolor{LightGreen}            & Adjective-Weight                                              & the larger ball is [lighter] and the smaller one is [heavier]                                                                 \\
 \rowcolor{LighterGreen}          & Verb-Intransitive, Noun                                       & the dog is [standing] and the person is [swimming]                                                                            \\
 \rowcolor{LightGreen}            & Verb-Intransitive Phrase, Adverb-Animate                      & the person on the left is [crying sadly] while the one on the right is [smiling happily]                                      \\
 \rowcolor{LighterGreen}          & Scope, Relative Clause                                        & a fencer [who is wearing black pants] having a point scored against them by another fencer [] using a wheelchair              \\
 \rowcolor{LightGreen}            & Adjective-Speed                                               & the train is [still] while the person is [moving fast]                                                                        \\
 \rowcolor{LighterGreen}          & Adverb-Temporal                                               & a person is drinking [now] and eating [later]                                                                                 \\
 \rowcolor{LightGreen}            & Adverb-Spatial                                                & the car is sitting [upside down] while the person is standing [rightside up]                                                  \\
 \rowcolor{LighterGreen} Relation & Adjective-Shape                                               & the [round] table has a [square] base                                                                                         \\
 \rowcolor{LightGreen}            & Noun, Adjective-Color                                         & Young person playing baseball with a [blue] bat and [green] ball                                                              \\
 \rowcolor{LighterGreen}          & Verb-Transitive                                               & the person with the ponytail [buys] stuff and other [packs] it                                                                \\
 \rowcolor{LightGreen}            & Scope, Verb-Transitive                                        & [] gears for [moving] something                                                                                               \\
 \rowcolor{LighterGreen}          & Scope, Preposition Phrase                                     & [] child in [front facing] row of yellow rubber ducks                                                                         \\
 \rowcolor{LightGreen}            & Adjective-Temperature                                         & a [hot] drink on a [cold] day                                                                                                 \\
 \rowcolor{LighterGreen}          & Adjective-Temporal                                            & the [first] vowel is E and the [last] consonant is N                                                                          \\
 \rowcolor{LightGreen}            & Scope, Conjunction                                            & a person spraying water on [someone else]1 [and]2 a person on a bike []2 []1                                                  \\
 \rowcolor{LighterGreen}          & Scope, Conjunction Phrase                                     & A child [] riding a bike [and an adult]                                                                                       \\
 \rowcolor{LightGreen}            & Preposition Phrase, Scope                                     & someone [with an apple] is hurt by a tree []                                                                                  \\
 \rowcolor{LighterGreen}          & Adjective-Manner Phrase                                       & two people wearing clothes of [different] colors are on [the same] side of the tennis net                                     \\
 \rowcolor{LightGreen}            & Verb-Intransitive                                             & a person [stands] and a dog [sits]                                                                                            \\
 \rowcolor{LighterGreen}          & Adjective-Animate                                             & [toy] cat with [real] baby                                                                                                    \\
 \rowcolor{LightGreen}            & Adverb-Spatial Phrase                                         & the sailboat sails [close] but the beach is [far away]                                                                        \\
 \rowcolor{LighterGreen}          & Scope, Adjective-Texture                                      & A [] small animal with [curled] hair                                                                                          \\
 \rowcolor{LightGreen}            & Adverb-Animate                                                & someone talks on the phone [angrily] while another person sits [happily]                                                      \\
 \rowcolor{LighterGreen}          & Adjective-Manner                                              & [poor] [unfortunate] people                                                                                                   \\
 \rowcolor{LightGreen}            & Verb-Transitive Phrase                                        & they [drank water] then they [worked out]                                                                                     \\
 \rowcolor{LighterGreen}          & Adjective-Color (3-way swap)                                  & The [red]$\rightarrow$[yellow] book is above the [yellow]$\rightarrow$[blue] book and below the [blue]$\rightarrow$[red] book       \\
 \rowcolor{LightGreen}            & Scope, Adjective-Manner                                       & [] living things [drinking]                                                                                                   \\
 \rowcolor{LighterGreen}          & Preposition                                                   & seat numbers increasing from [right] to [left]                                                                                \\
 \rowcolor{LightGreen}            & Verb-Intransitive Phrase                                      & a cat is [stretching] and a person is [lying down]                                                                            \\
 \rowcolor{LighterGreen}          & Sentence                                                      & [the coffee is poured] before [it is ground]                                                                                  \\
 \rowcolor{LightGreen}            & Adjective-Speed Phrase, Verb-Intransitive                     & the person with green legs is running [quite slowly] and the red legged one runs [faster]                                     \\
 \rowcolor{LighterGreen}          & Adjective-Spatial                                             & A [left] hand pulls a glove onto a [right] hand                                                                               \\
 \rowcolor{LightGreen}            & Negation, Scope                                               & The [un]caged bird has an []opened cage door                                                                                  \\
 \rowcolor{LighterGreen}          & Verb-Transitive Phrase, Verb-Intransitive, Preposition Phrase & the dog [bite]1s []2 what someone would normally [wear]1 [as a hat]2                                                          \\
 \rowcolor{LightYellow} \midrule    & Altered POS                                                   & [watch]ing the [present]                                                                                                      \\
 \rowcolor{LighterYellow}         & Verb-Transitive, Noun                                         & someone []1 on [the ground]2 [is]1 spraying water towards [a vehicle]2                                                        \\
 \rowcolor{LightYellow}           & Scope, Altered POS, Verb-Intransitive, Verb-Transitive        & [walking]1 someone []1 [cut]2 [lines]2 into green plants                                                                      \\
 \rowcolor{LighterYellow}    & Noun, Adjective-Size                                          & the [person]1 is too [big]2 for the [small]2 [door]1                                                                          \\
 \rowcolor{LightYellow} Both           & Noun, Verb-Intransitive                                       & a [dog sitting] on a couch with a [person lying] on the floor                                                                 \\
 \rowcolor{LighterYellow}         & Scope, Noun, Preposition                                      & []1 a person [near]1 [water]2 using a []2 lasso                                                                               \\
 \rowcolor{LightYellow}           & Noun, Preposition Phrase, Scope                               & a person wearing a [bear]1 mask []2 in blue on the left hand side of a person wearing a [panda]1 mask [with glasses]2 in pink \\
 \rowcolor{LighterYellow}         & Scope, Preposition Phrase, Adjective-Color                    & [darker]1 things []2 become [light]1 [in stripes]2                                                                            \\
 \rowcolor{LightYellow}           & Altered POS, Determiner-Numeral                               & [one] ear that some [donkey] is whispering a secret into                                                                      \\
\bottomrule
\end{tabular}
}
\caption{Examples showcasing the full linguistic (swap-dependent) tag breakdown.}
\end{table}
\newpage

\section{Heatmaps for the Word-Region Alignment Models}

We provide heatmaps for models that were trained with a word-region alignment objective: UNITER, ViLLA and ViLT. See the main text for ViLT heatmaps.

\begin{figure}[H]
    \centering
    \begin{minipage}{0.48\textwidth}
        \centering
        \scalebox{0.9}{
            \begin{tabular}{cccc}
            \multicolumn{3}{c}{a \textbf{brown dog} is on a \textbf{white couch}}\\
             \includegraphics[height=1.5cm, width=0.25\textwidth]{vilt_ot_heatmap/e29_i0_base.png}& \includegraphics[height=1.5cm, width=0.25\textwidth]{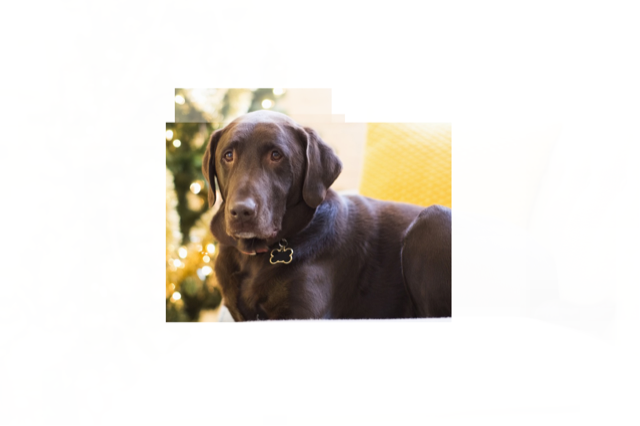} & \includegraphics[height=1.5cm, width=0.25\textwidth]{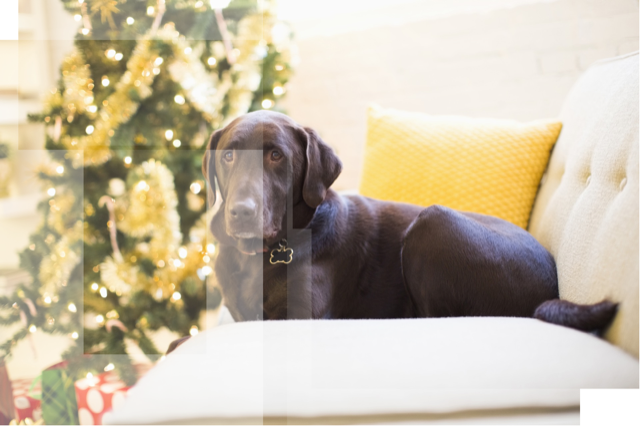} \\
             \includegraphics[height=1.5cm, width=0.25\textwidth]{vilt_ot_heatmap/e29_i1_base.png}& \includegraphics[height=1.5cm, width=0.25\textwidth]{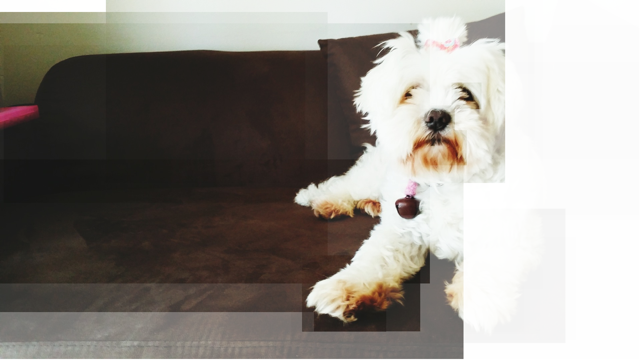} & \includegraphics[height=1.5cm, width=0.25\textwidth]{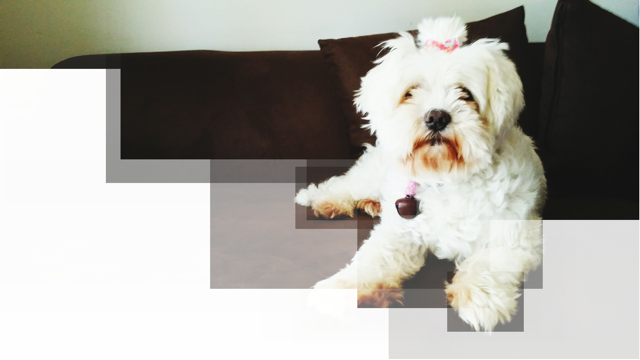} \\
             & & & \\
             \multicolumn{3}{c}{a \textbf{white dog}\quad is on a \quad\textbf{brown couch}}\\
             \includegraphics[height=1.5cm, width=0.25\textwidth]{vilt_ot_heatmap/e29_i1_base.png}& \includegraphics[height=1.5cm, width=0.25\textwidth]{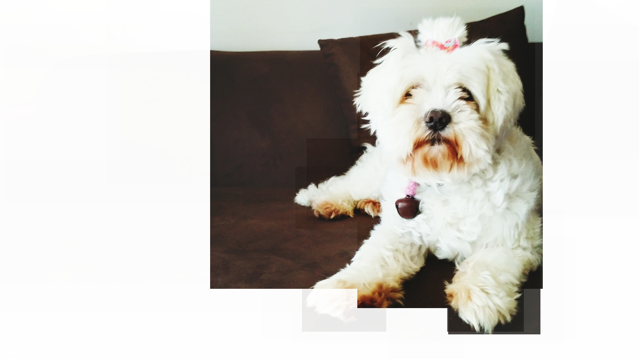} & \includegraphics[height=1.5cm, width=0.25\textwidth]{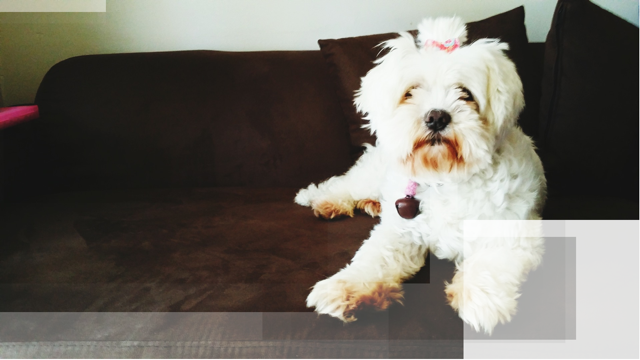} \\
             \includegraphics[height=1.5cm, width=0.25\textwidth]{vilt_ot_heatmap/e29_i0_base.png}& \includegraphics[height=1.5cm, width=0.25\textwidth]{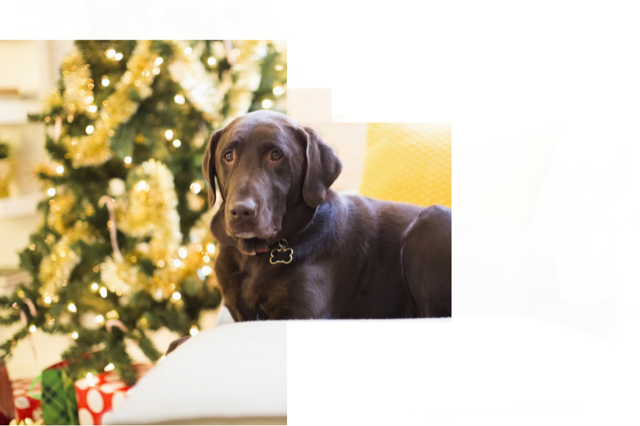} & \includegraphics[height=1.5cm, width=0.25\textwidth]{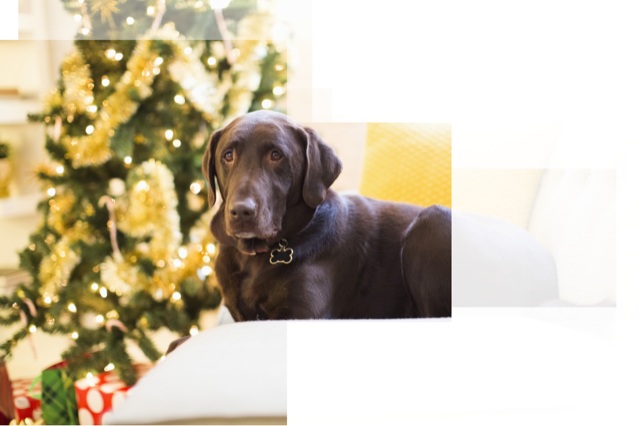} \\
            \end{tabular}
        }
    \end{minipage}
    ~
    \begin{minipage}{0.48\textwidth}
        \centering
        \scalebox{0.9}{
            \begin{tabular}{cccc}
            \multicolumn{3}{c}{\textbf{circular food} on \textbf{heart-shaped wood}}\\
            \includegraphics[height=1.5cm, width=0.20\textwidth]{vilt_ot_heatmap/e118_img_0.png}& \includegraphics[height=1.5cm, width=0.20\textwidth]{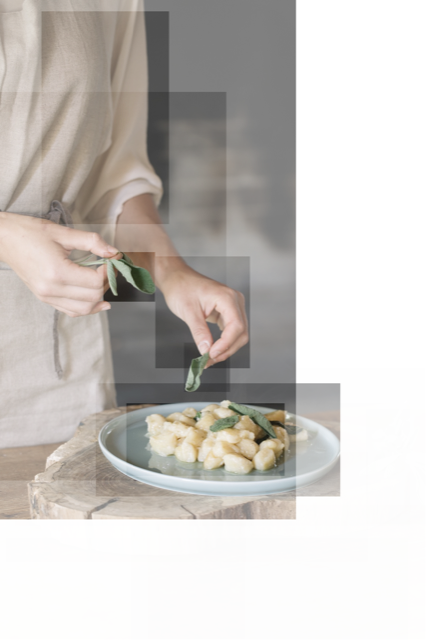} & \includegraphics[height=1.5cm, width=0.20\textwidth]{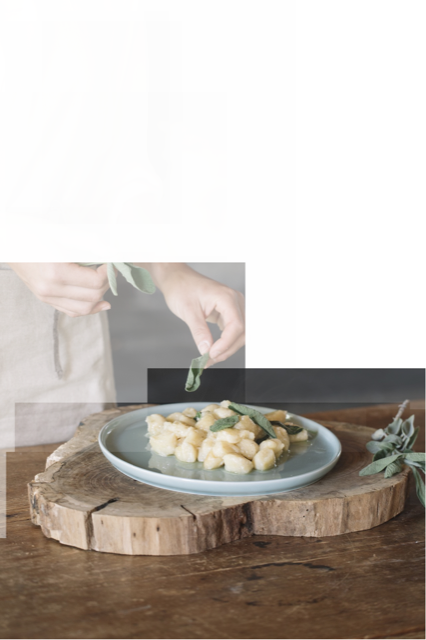} \\
            \includegraphics[height=1.5cm, width=0.25\textwidth]{vilt_ot_heatmap/e118_img_1.png}& \includegraphics[height=1.5cm, width=0.25\textwidth]{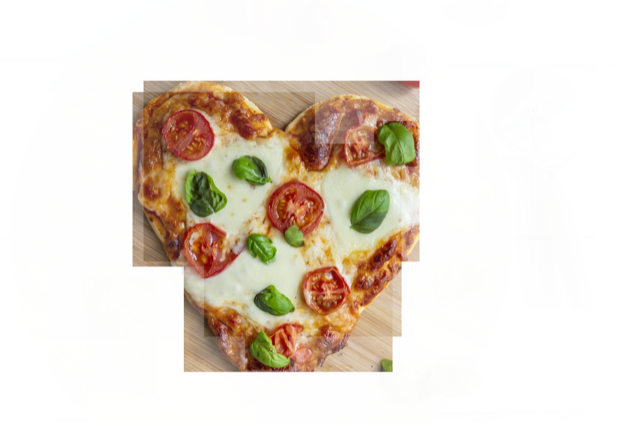} & \includegraphics[height=1.5cm, width=0.25\textwidth]{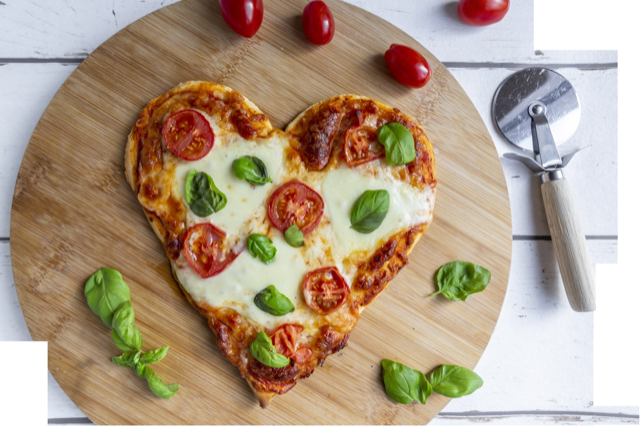} \\
             & & & \\
             \multicolumn{3}{c}{ \textbf{heart-shaped food} on \textbf{circular wood}}\\
            \includegraphics[height=1.5cm, width=0.25\textwidth]{vilt_ot_heatmap/e118_img_1.png}& \includegraphics[height=1.5cm, width=0.25\textwidth]{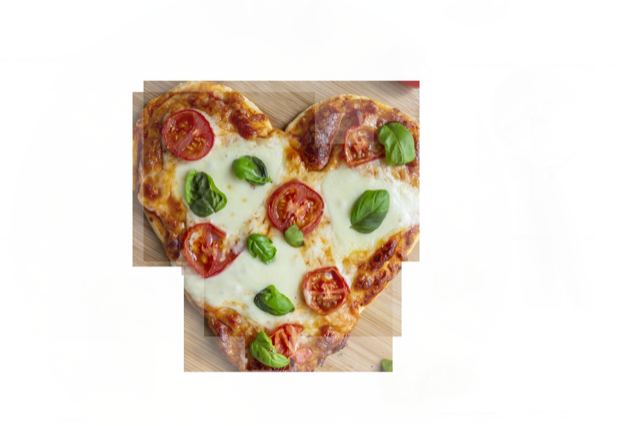} & \includegraphics[height=1.5cm, width=0.25\textwidth]{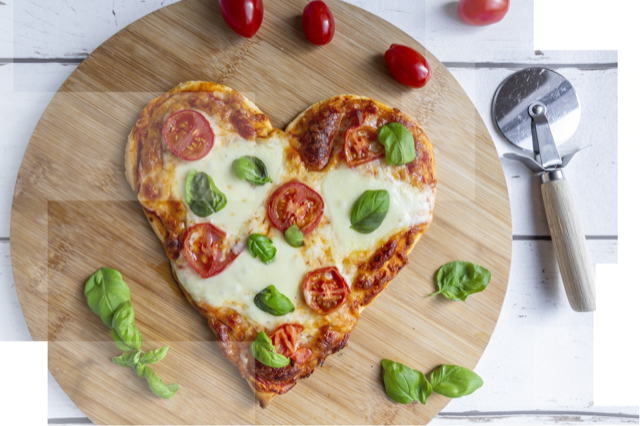} \\
            \includegraphics[height=1.5cm, width=0.20\textwidth]{vilt_ot_heatmap/e118_img_0.png}& \includegraphics[height=1.5cm, width=0.20\textwidth]{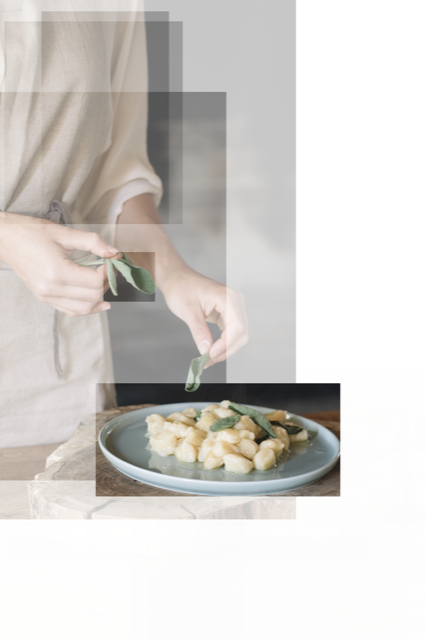} & \includegraphics[height=1.5cm, width=0.20\textwidth]{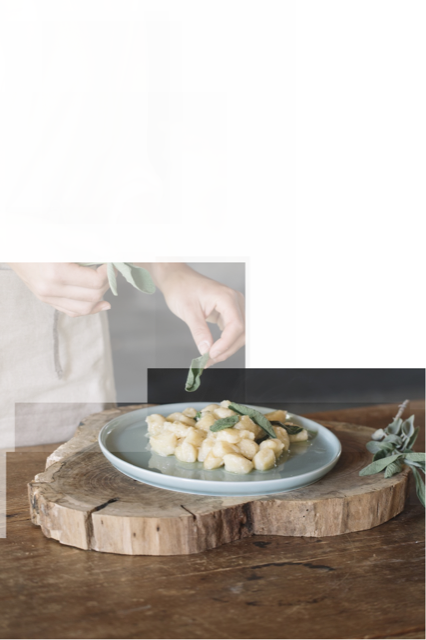} \\
            \end{tabular}
        }
    \end{minipage}
    
    \caption{Word-region alignment scores between the image and text features for ViLLA$_{base}$ on examples from Winoground.}
    \label{fig:villa-base-heatmap}
\end{figure}

\begin{figure}[H]
    \centering
    \begin{minipage}{0.48\textwidth}
        \centering
        \scalebox{0.9}{
            \begin{tabular}{cccc}
            \multicolumn{3}{c}{a \textbf{brown dog} is on a \textbf{white couch}}\\
             \includegraphics[height=1.5cm, width=0.25\textwidth]{vilt_ot_heatmap/e29_i0_base.png}& \includegraphics[height=1.5cm, width=0.25\textwidth]{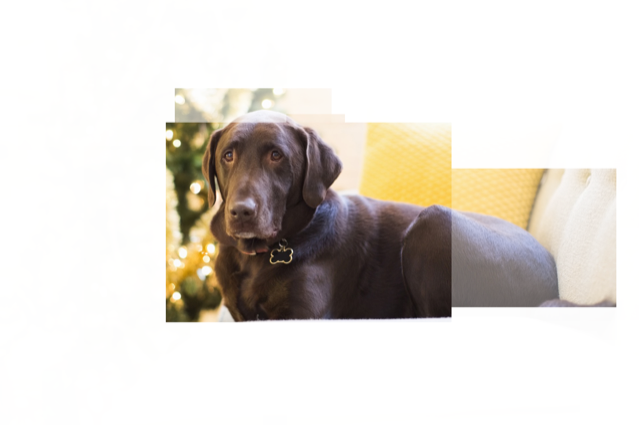} & \includegraphics[height=1.5cm, width=0.25\textwidth]{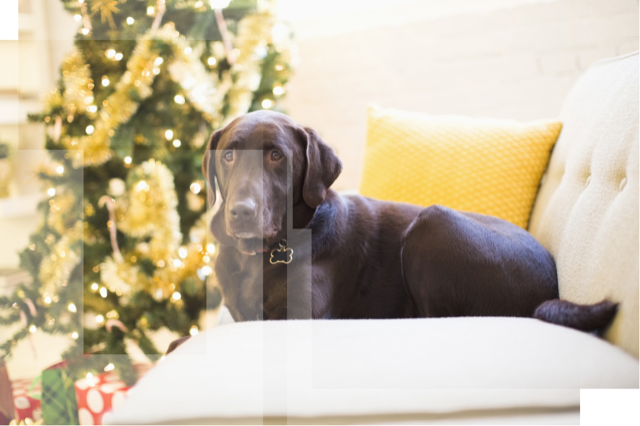} \\
             \includegraphics[height=1.5cm, width=0.25\textwidth]{vilt_ot_heatmap/e29_i1_base.png}& \includegraphics[height=1.5cm, width=0.25\textwidth]{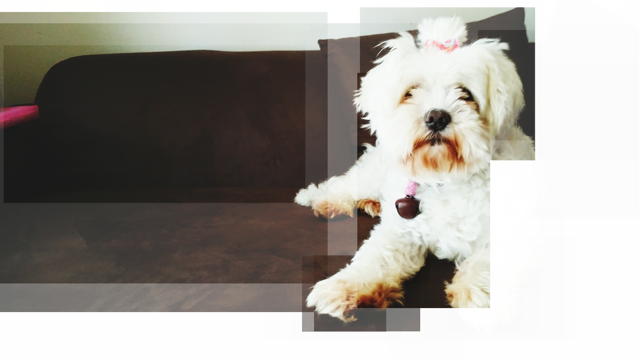} & \includegraphics[height=1.5cm, width=0.25\textwidth]{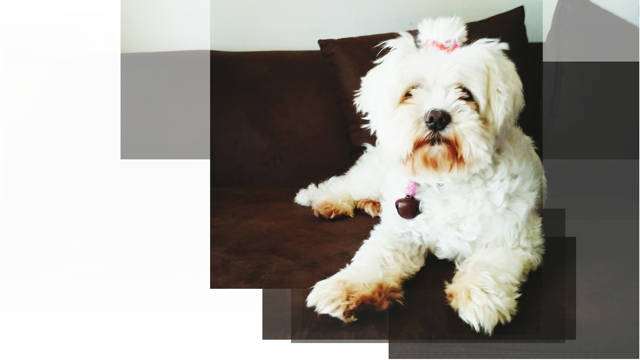} \\
             & & & \\
             \multicolumn{3}{c}{a \textbf{white dog}\quad is on a \quad\textbf{brown couch}}\\
             \includegraphics[height=1.5cm, width=0.25\textwidth]{vilt_ot_heatmap/e29_i1_base.png}& \includegraphics[height=1.5cm, width=0.25\textwidth]{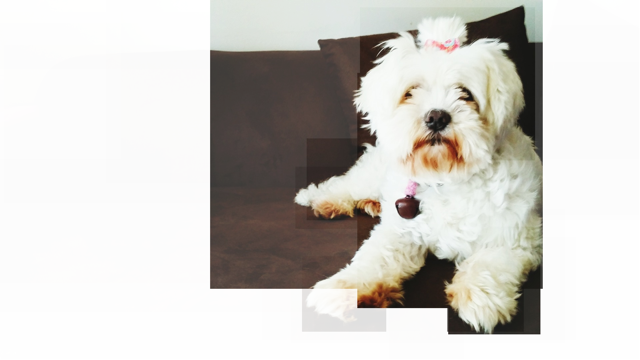} & \includegraphics[height=1.5cm, width=0.25\textwidth]{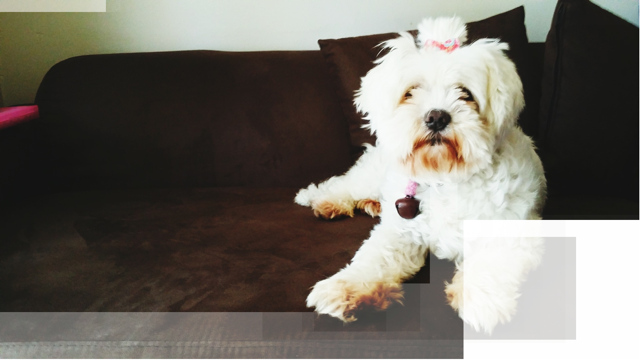} \\
             \includegraphics[height=1.5cm, width=0.25\textwidth]{vilt_ot_heatmap/e29_i0_base.png}& \includegraphics[height=1.5cm, width=0.25\textwidth]{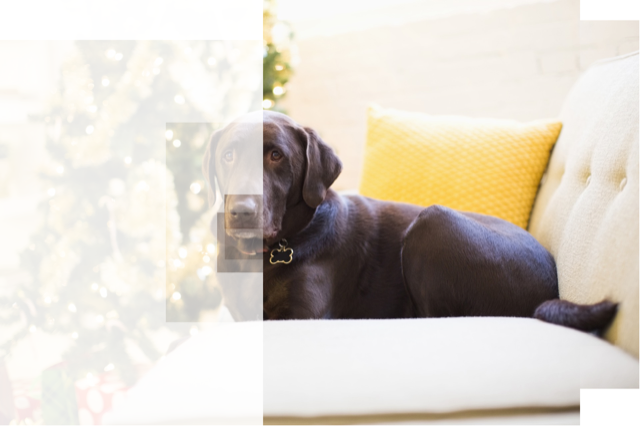} & \includegraphics[height=1.5cm, width=0.25\textwidth]{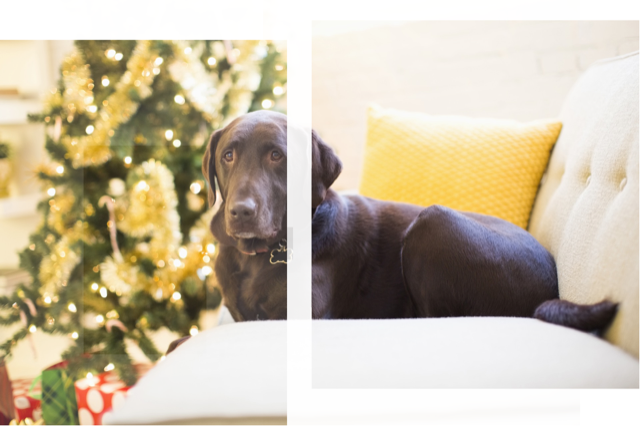} \\
            \end{tabular}
        }
    \end{minipage}
    ~
    \begin{minipage}{0.48\textwidth}
        \centering
        \scalebox{0.9}{
            \begin{tabular}{cccc}
            \multicolumn{3}{c}{\textbf{circular food} on \textbf{heart-shaped wood}}\\
            \includegraphics[height=1.5cm, width=0.20\textwidth]{vilt_ot_heatmap/e118_img_0.png}& \includegraphics[height=1.5cm, width=0.20\textwidth]{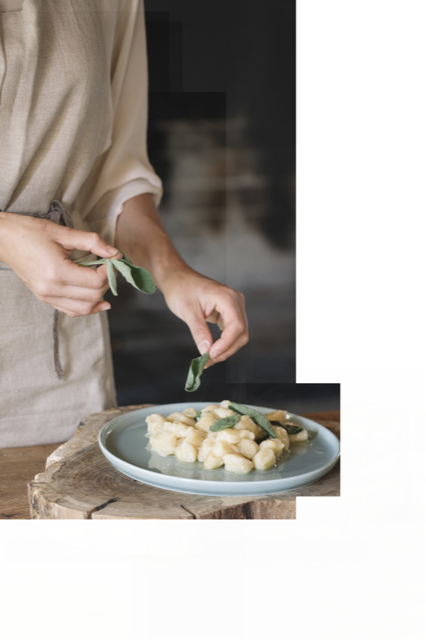} & \includegraphics[height=1.5cm, width=0.20\textwidth]{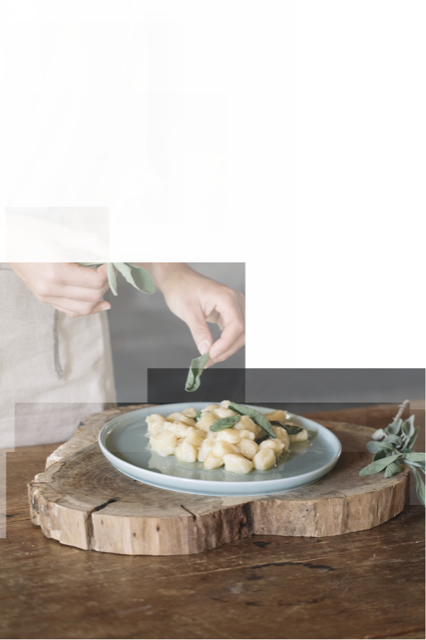} \\
            \includegraphics[height=1.5cm, width=0.25\textwidth]{vilt_ot_heatmap/e118_img_1.png}& \includegraphics[height=1.5cm, width=0.25\textwidth]{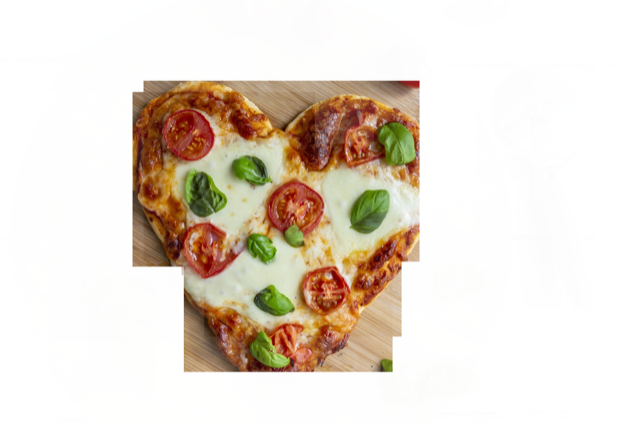} & \includegraphics[height=1.5cm, width=0.25\textwidth]{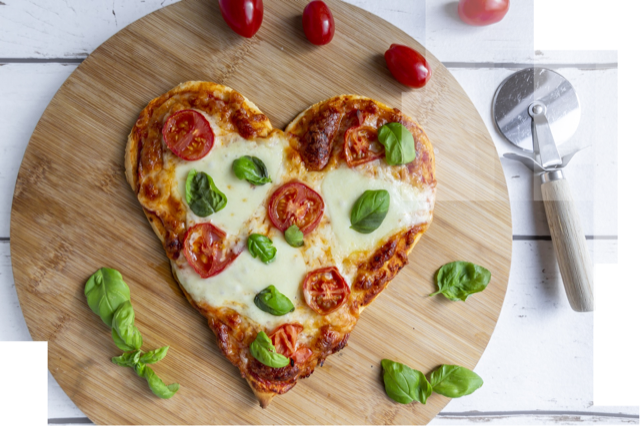} \\
             & & & \\
             \multicolumn{3}{c}{ \textbf{heart-shaped food} on \textbf{circular wood}}\\
            \includegraphics[height=1.5cm, width=0.25\textwidth]{vilt_ot_heatmap/e118_img_1.png}& \includegraphics[height=1.5cm, width=0.25\textwidth]{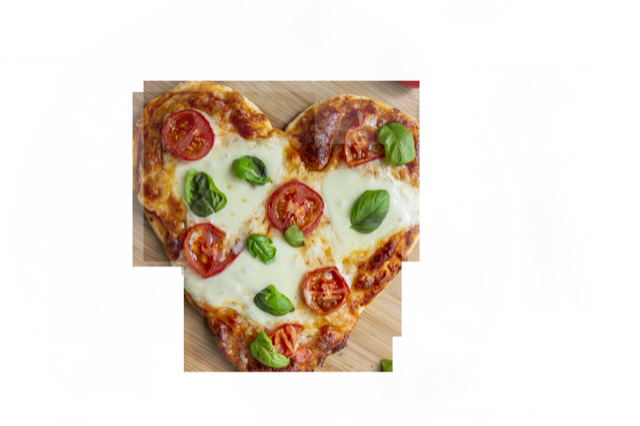} & \includegraphics[height=1.5cm, width=0.25\textwidth]{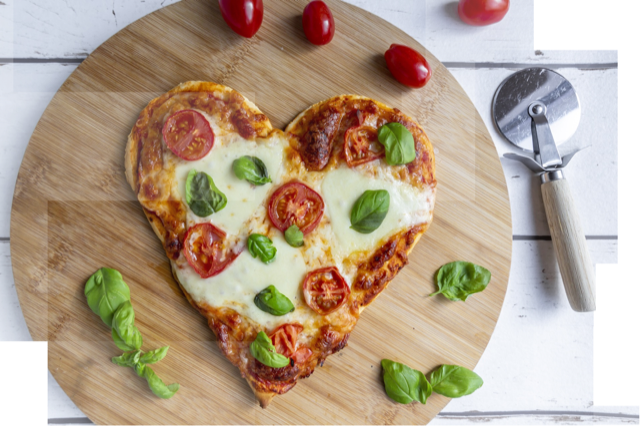} \\
            \includegraphics[height=1.5cm, width=0.20\textwidth]{vilt_ot_heatmap/e118_img_0.png}& \includegraphics[height=1.5cm, width=0.20\textwidth]{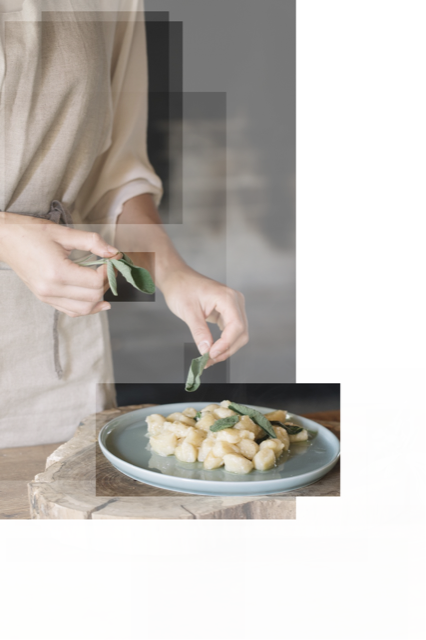} & \includegraphics[height=1.5cm, width=0.20\textwidth]{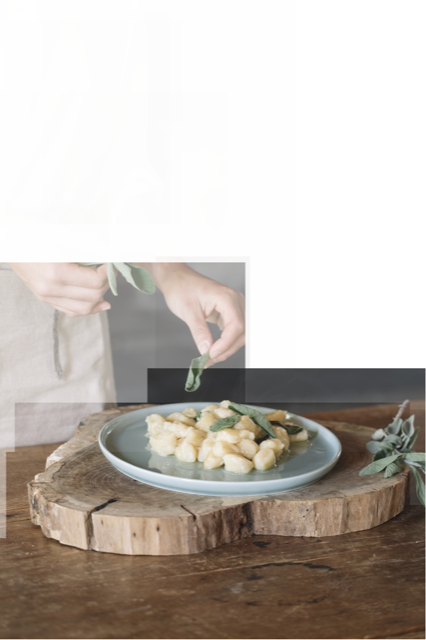} \\
            \end{tabular}
        }
    \end{minipage}
    
    \caption{Word-region alignment scores between the image and text features for UNITER$_{base}$ on examples from Winoground.}
    \label{fig:uniter-base-heatmap}
\end{figure}

\newpage

\section{Mechanical Turk Interface}

In order to participate, crowdworkers needed to satisfy several criteria: be an English speaker, have 98\% previous HIT approval, have completed 1000 previous HITs, and pass the onboarding test. The onboarding test used the same interface as the actual task. It consisted of ten image-caption match questions, with images and captions that are independent from the actual Winoground dataset. If they made one mistake, a pop-up asked them if they were sure, and they would be allowed to select whether there was a match or not again. If they made any additional mistakes during onboarding, they were disqualified.

\begin{figure}[H]\label{fig:amt}
    \centering
    \includegraphics[width=\textwidth]{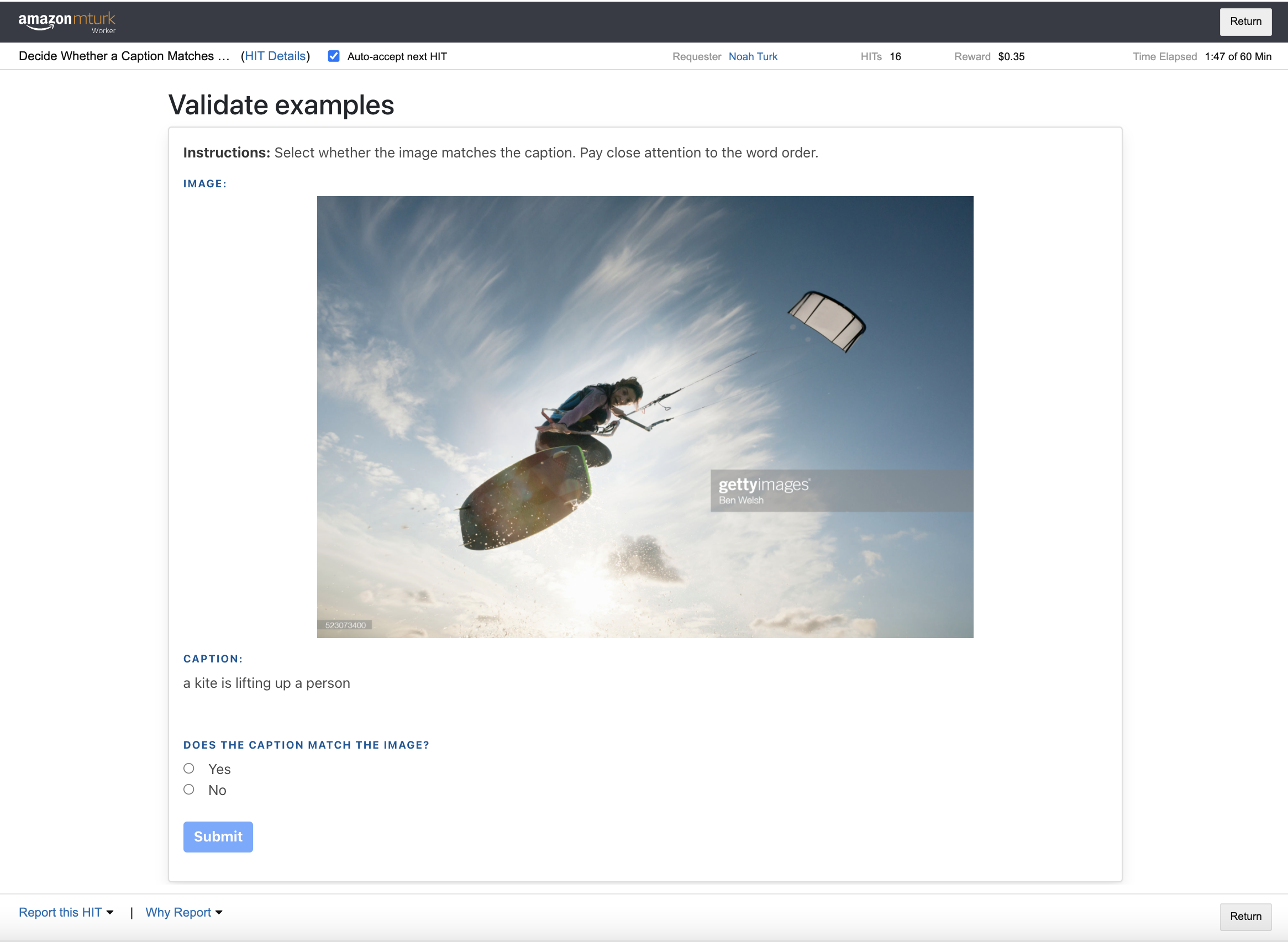}
    \caption{The Amazon Mechanical Turk validation interface.}
\end{figure}

\newpage

\section{Ethical Considerations}

A key consideration while designing Winoground centered on how the expert annotators would describe the people contained in the images.
We avoided using gendered terms (\eg using ``person" in place of ``woman" or ``man") in our captions and did not include any swaps between pairs of captions based on gender, race or ethnicity (\eg ``\textit{[the man] hands a water to [the woman]"}).
We recognize that, barring direct access to the people in the images, we would be merely making a guess at a person's identity based on our own cultural norms and experiences.

In addition, we encouraged the expert annotators to find images that represent a variety of people across the dimensions of perceived race, gender, disability, \textit{etc.}.
We gathered the Getty Images metadata (title and short alt text-like description) and searched them for specific words as a rough proxy for gender representation.
The relevant words are either words referring to women (\eg girl, her), words referring to men (\eg boy, him) or words that are gender-neutral (\eg them, themself).
Using the Getty Images metadata corresponding to the 800 images in Winoground, 371 images have corresponding metadata that contained at least one word from the lists we created.
Using this metadata for these 371 images, we estimate that 152 images only contain women, 123 images only contain men, 22 images only contain people without gender descriptors, and the remaining 74 images contain people described by multiple genders.
This serves only as a rough estimate as much of the metadata contain words referring to people that are inherently non-gendered (\eg scuba diver, friend, \etc) and because the relevant gendered words we found are themselves subject to the assumptions of those who wrote the titles and captions.